\documentclass{article}
\usepackage{spconf,amsmath,graphicx}
\usepackage{url}
\usepackage{booktabs} 
\usepackage{times}
\usepackage{mathptmx}
\usepackage[T1]{fontenc}
\usepackage[utf8]{inputenc}
\usepackage{hyperref}
\def\x{{\mathbf x}}
\def\L{{\cal L}}

\title{BRI3L: A brightness illusion image dataset for identification and localization of regions of illusory perception}
\name{Aniket Roy$^{1}$, Anirban Roy$^{2}$, Soma Mitra$^{3}$, Kuntal Ghosh$^{4}$ }
\address{$^{1}$Johns Hopkins University, $^{2}$SRI International, $^{3}$CDAC Kolkata, $^{4}$Indian Statistical Institute}

\begin{document}
%\ninept
%
\maketitle
\begin{abstract}
Visual illusions play a significant role in understanding visual perception. Current methods in understanding and evaluating visual illusions are mostly deterministic filtering based approach and they evaluate on a handful of visual illusions, and the conclusions therefore,  are not generic.
% The effectiveness of the visual cortex inspired deep learning models to understand and interpret visual illusions has  suffered due to lack of large-scale benchmark dataset.
To this end, we generate a large-scale dataset of 22,366 images (BRI3L: BRightness Illusion Image dataset for Identification and Localization of illusory perception) of the five types of brightness illusions and benchmark the dataset using data-driven neural network based approaches. The dataset contains label information - (1) whether a particular image is illusory/non-illusory, (2) the segmentation mask of the illusory region of the image. Hence, both the classification and segmentation task can be evaluated using this dataset. We follow the standard psychophysical experiments involving human subjects to validate the dataset.
To the best of our knowledge, this is the first attempt to develop a dataset of visual illusions and benchmark using data-driven approach for illusion classification and localization. 
% In this study, we consider  brightness illusions which are important for understanding the low and mid-level vision in particular, and where the perceived brightness of a region is different from the actual luminosity in the image. 
We consider five well-studied types of brightness illusions: 1) Hermann grid, 2) Simultaneous Brightness Contrast, 3) White illusion, 4) Grid illusion, and 5) Induced Grating illusion. 
Benchmarking on the dataset achieves 99.56\% accuracy in illusion identification and 84.37\% pixel accuracy in illusion localization. The application of deep learning model, it is shown, also generalizes over unseen brightness illusions like brightness assimilation to contrast transitions. We also test the ability of state-of-the-art diffusion models to generate brightness illusions. We have provided all the code, dataset, instructions etc in the github repo: \href{https://github.com/aniket004/BRI3L}{https://github.com/aniket004/BRI3L}
\end{abstract}
%

%%%%%%%%%%%%%%%%%%%%%%%%%%%%%%%%%%%%%%%%%%%%%%%%%%%%%%%%%%%%%%%%%%%%%%%%%%%%%%%%
\begin{figure}[!t]
	\centering
	\includegraphics[scale=0.23]{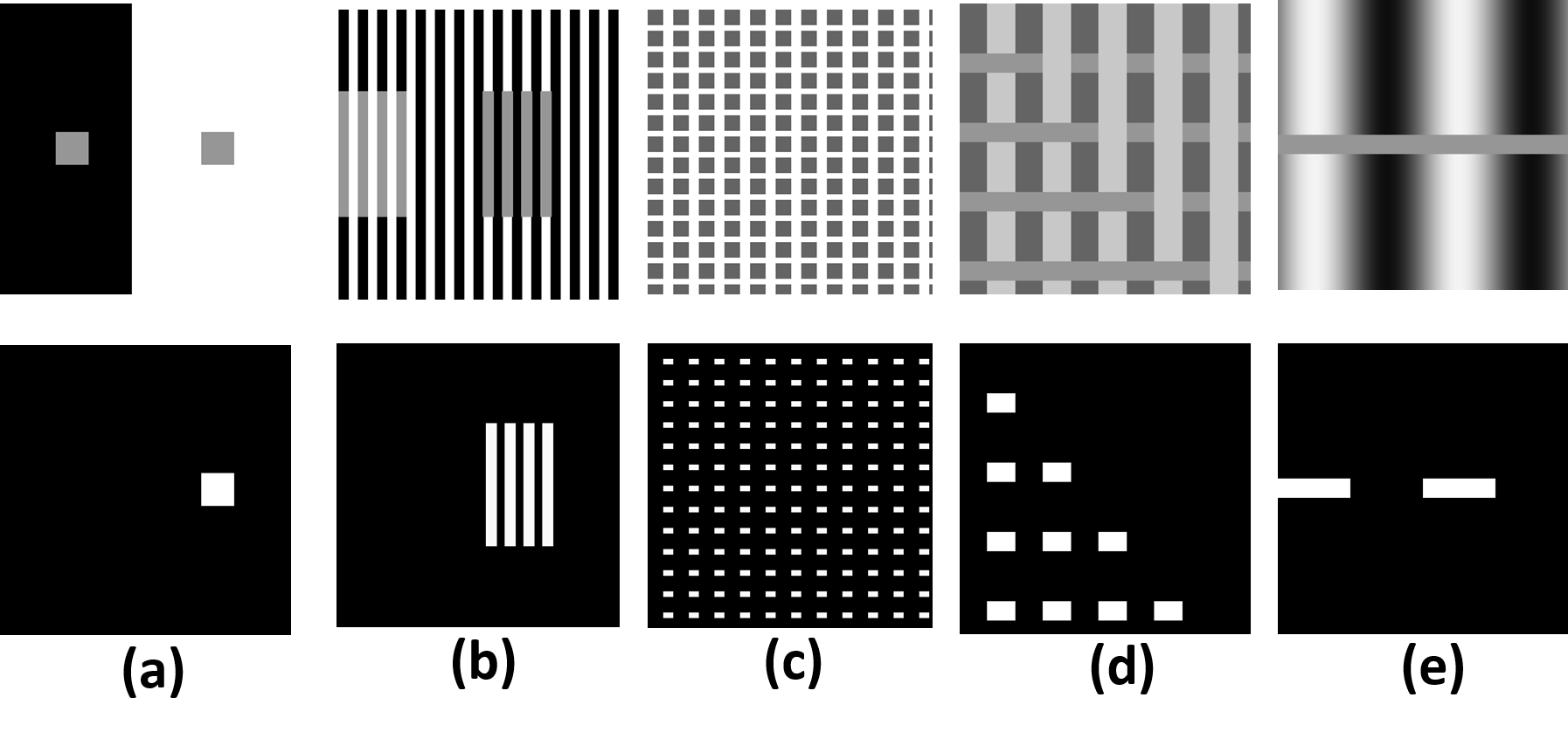}
	\vspace{-0.2cm}
	\caption{Examples of brightness illusion (top row) and the binary masks corresponding to certain illusory regions in the image (bottom row): (a) simultaneous brightness contrast (SBC), (b) White illusion, (c) Hermann grid, (d) grid illusion, (e) grating illusion. The illusory regions correspond to the illusory phenomenon such as apparent false perception of brightness/darkness values based on the context. Our goal is to localize these illusory regions in the illusory images.}
	\vspace{-0.6cm}
	\label{fig:illusion}
\end{figure}
%%%%%%%%%%%%%%%%%%%%%%%%%%%%%%%%%%%%%%%%%%%%%%%%%%%%%%%%%%%%%%%%%%%%%%%%%%%%%%%%%%%

\section{Introduction}
\label{sec:introduction}

\vspace{-0.2cm}
% Recent studies in computational neuroscience have shown that analyzing visual illusions might provide insight towards similarities and differences between biological and machine vision~\cite{gomez2020color, gomez2019synthesizing, kim2020neural, benjamin2019shared}. 
% Machine vision has been inspired from biological vision, which get insights from psychophysical data like illusions~\cite{gomez2020color, gomez2019synthesizing}. With the advances of deep learning, itself inspired by visual cortical modeling, we have powerful neural network based tools to analyse psychophysical data to get a better understanding of biological vision. 
% However, these deep neural network models are data-hungry and lack of psychophysical data, e.g., illusions hinders the progress in understanding and evaluating the intriguing phenomenon of illusion perception. To this end, we generate a large-scale dataset consisting of brightness illusions.

Brightness illusions are a special and well explored class of visual illusions where the perceived brightness of a region is different from the actual brightness in the image. A few instances of the brightness illusion and the corresponding illusory regions are shown in Fig.~\ref{fig:illusion}. Brightness illusions can be classified into three major classes according to changes in directions of the brightness: 1) brightness-contrast, 2) brightness-assimilation, and 3) illusory blobs or region illusions~\cite{ghosh2012possible}. In brightness-contrast illusions, the brightness induction in the test patch occurs in opposite direction to the surrounding luminance such as the simultaneous brightness contrast (SBC)~\cite{heinemann1955simultaneous} in Fig.~\ref{fig:illusion} (a). In this example, the right test patch looks darker since it is surrounded by brighter background causing a contrast of brightness.
In brightness-assimilation illusions, the brightness induction occurs in the same direction such as the White illusion~\cite{white1979new} in Fig.~\ref{fig:illusion} (b). In this example, the right test patches look darker while it is surrounded by darker background causing an assimilation of brightness.
In the illusory blobs or region illusions, pseudo bright or dark blobs, bands, or contours appear in the image corners such as the Hermann grid~\cite{Hermann1870erscheinung} as shown in Fig.~\ref{fig:illusion} (c). Some other complex brightness perception illusions like the grid illusion (Fig.~\ref{fig:illusion} (d)) and the grating illusion~\cite{foley1985visual} (Fig.~\ref{fig:illusion} (e)) are also shown.

Localizing illusory regions is important as it helps to identify the nature of illusions and facilitate understanding the reason behind the illusory phenomenon \cite{kingdom2011lightness,kriegeskorte2015deep, ghosh2012possible}. Localizing illusory regions in a brightness illusion is challenging as the intensity values in the illusory and non-illusory regions can be exactly the same (Fig.~\ref{fig:illusion}(a) top). Though these regions appear to be of different brightness values to humans, it can be difficult for image processing approaches to differentiate them. For example, various threshold-based illusion segmentation approaches, e.g., Otsu \cite{otsu1979threshold}, can not differentiate the illusory regions from the non-illusory regions as the intensity values are the same in those regions.

Traditionally, spatial filtering based approaches are employed to localize illusions in images~\cite{blakeslee2004unified,robinson2007filtering,ghosh2012possible} where hand-crafted filters, such as the Difference of Gaussians (DOG), Laplacian of a Gaussian (LOG), and their more contemporary versions (ODOG, LODOG, FLODOG) ~\cite{robinson2007filtering} are designed to localize illusory regions in the images. However, as noted in some papers \cite{gilchrist1999anchoring,kingdom2011lightness}, these filtering based approaches are inadequate in explaining complex versions of brightness illusions. Gilchrist et al.~\cite{gilchrist1999anchoring} refer the spatial filtering based models as intrinsic image based models and suggest an alternative approach for brightness perception based on anchoring theory. Recently, Gomez et. al.~\cite{gomez2018convolutional} show that training a shallow neural network on low-level tasks (e.g., image denoising) can be effective to explain visual illusions. However, their approach does not generalize across various gray-scale illusions such as the higher spatial frequency induced assimilation illusions~\cite{gomez2018convolutional}.

To address the above-mentioned challenges, we introduce a large-scale dataset of five types of commonly used brightness illusion with the annotated illusion regions (Fig.~\ref{fig:illusion} (bottom row)). Then we benchmark the dataset (BRI3L: BRightness Illusion Image dataset for Identification and Localization of illusory perception)
using a data-driven neural-network approach to distinguishing illusions from non-illusions and localize the illusory regions in the illusion images. Specifically, given an image, we identify if the image represents an illusion and localize the illusory regions in the image. We consider a ResNet (a neural network with residual connections) \cite{he2016deep} to classify the illusion images from non-illusion images. For illusion localization, we consider a U-Net (an encoder-decoder neural network) \cite{ronneberger2015u} to segment the illusory regions in illusion images. Moreover, we show that neural networks trained on a specific type of illusions, can generalize to unseen types of illusions.

Our main contributions include:
%\begin{itemize}[noitemsep, topsep=0pt,leftmargin=*]
\begin{itemize}
\setlength\itemsep{0em}
    \item We introduce a large-scale dataset of 22,366 images containing five types of brightness illusions. We annotate each image with the binary mask corresponding to illusory regions in the image. We validate the dataset by conducting standard psychophysical experiments involving human experts. The dataset and training details are provided in the supplementary material.
    \item A data-driven neural network-based approach to identify brightness illusions and localize the illusory regions. To the best of our knowledge, this is the first attempt to consider a data-driven approach for illusion localization.
    \item Unlike the common filtering based approaches, our approach is generalizable to novel types of illusions. Moreover, we can also address the intriguing transition from brightness contrast to assimilation. 
    \item We consider a perception-inspired structural similarity loss suitable for illusion localization. Our experiments show that this loss is crucial for illusion localization.
    \item This work reports the first attempt to investigate the ability of state-of-the-art diffusion models to generate brightness illusions.
\end{itemize}

\vspace{-0.2cm}
\section{Prior work}
\label{sec:related_work}
\vspace{-0.2cm}

The approaches for illusion analysis can be broadly classified into two categorized as follows.

\textbf{Filtering based approaches.} There exists a plethora of filtering based approaches for illusion understanding that are mainly inspired by the visual pathway and brain structure and try to establish a one-to-one mapping with some of the biological structure of the visual system to explain several classes of brightness illusion \cite{whittle1992brightness,blakeslee2004unified,robinson2007filtering,ghosh2012possible,laparra2015visual,spitzer2005computational}. For example, the Oriented Difference of Gaussian (ODOG) model suggests that the input image is filtered with several directional Difference of Gaussian (DOG) filters which is mostly inspired by the Hubel-Wiesel's experiments in V1 cat and macaque cells~\cite{kandel2000principles}. Frequency-specific Locally normalized ODOG (FLODOG) ~\cite{robinson2007filtering} further suggests that the normalization step in ODOG model should be performed locally and in frequency sensitive manner. Similar strategies has been adopted by~\cite{ghosh2012possible, otazu2010toward,otazu2008multiresolution}.

% , which is more biologically plausible than ODOG, and consequently, more brightness illusions could be explained with this model. 
% Ghosh et al.~\cite{ghosh2012possible} modified the classical DOG model by incorporating an additional additive Gaussian filter as an Extended Center Receptive Field (ECRF), suggesting a resemblance with the Koniocellur cells in human perception. Otazu et al.~\cite{otazu2010toward,otazu2008multiresolution} predicted brightness induction effects by a unified chromatic induction model based on multiresolution wavelet transform. The unsupervised approach using multi-scale wavelet-based edge and texture detection with shallow networks has been proposed to locate illusory contours as well~\cite{manjunath1993unified}.

\textbf{Data-driven approaches.} In spite of the tremendous success of neural networks in various vision application, some recent studies show that they may not be able to emulate some basic perceptual phenomenon~\cite{martinez2019praise}. Therefore, recently a line of research devoted to studying the similarities and differences of biological and machine vision. Kim et al.~\cite{kim2020neural} showed that neural networks trained with natural images for classification exhibit the Gestalt's law of closure. Benjamin et al.~\cite{benjamin2019shared} observed that DNNs trained on Imagenet for classification shows orientation bias for geometric illusions similar to human observers. Ward et al.~\cite{ward2019exploring} verified that DNN trained for object recognition exhibits geometric Muller-Lyon illusion. Serre et al.~\cite{linsley2020recurrent} designed a deep RNN inspired by the orientation-tilt illusion, which outperforms state-of-the-art contour detection approach. Recently, Watanabe et al.~\cite{watanabe2018illusory} train a self-supervised predictive network (Pred-Net) for video prediction and apply it to successfully predict the illusory motion of the rotating snake illusion. Gomez et. al.~\cite{gomez2018convolutional} show that a network on low-level tasks, such as denoising, deblurring, can explain the response of visual illusions. In a subsequent work~\cite{gomez2019synthesizing}, a GAN is employed to generate visual illusions by constructing background for a target with aid of an illusion discriminator. Williams et al.~\cite{williams2018optical} also consider GANs for generating visual illusions, however, they are unable to fool human vision. Recently, Hirsch et al.~\cite{hirsch2020color} also used flow-based methods to generate color visual images.

\vspace{-0.2cm}
\section{BRI3L: A Brightness illusion image dataset}

\vspace{-0.2cm}
\subsection{Dataset generation}
\vspace{-0.2cm}

To facilitate research on illusion understanding, we create a large-scale dataset containing five types of brightness illusions: 1) Hermann grid, 2) SBC, 3) White illusion, 4) grating, 5) Grid illusion and non-illusion variants.

\textbf{Illusion images}
We synthetically generate illusion images for each of these five classes of brightness illusions as shown in Fig.~\ref{fig:illusion}.
Each image is annotated with the binary segmentation mask corresponding to illusory regions in the image. The dataset is created by considering several variations of illusion-related parameters including patch and grid width, height, and intensity values. The dataset contains 22366 illusion images of size 256$\times$256 consisting of 4160 SBC illusions, 637 White illusions, 1024 Hermann grid illusions, 6350 induced grating illusions, and 10195 upper and lower grid illusions. The grayscale values are varied in ranges \{100, 150, 200, 250\} grayscale values, and the aspect ratio ranges of \{0.2, 0.4, 0.8\} for SBC and White illusions. The spatial frequency of grating illusions vary from 4-50 cycles per degrees. More examples are provided in the supplementary material.

\begin{figure}[!t]
\centering
\includegraphics[scale=0.4]{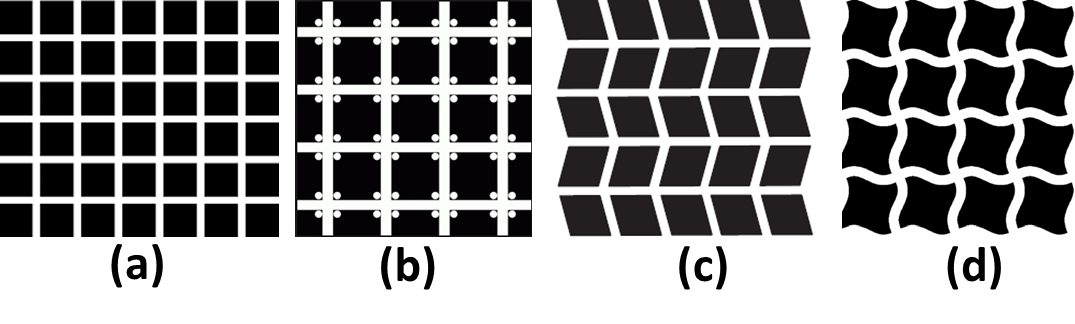}
\vspace{-0.3cm}
\caption{(a): Actual Hermann grid; Weak or non-illusion variants of Hermann grid by (b) inserting blobs, (c) and (d) introducing non-linearity.~\cite{geier2008straightness}}
\vspace{-0.6cm}
\label{fig:non_linear}
\end{figure}

\textbf{Non-illusion images}
To distinguish between illusion vs non-illusion, non-illusion samples are also required. We generate 1149 non-illusion images that are carefully created by slightly modifying the image to make the illusory effect disappear~\cite{geier2008straightness,bakshi2020tiny}. Some of the non-illusions are shown in Fig.~\ref{fig:non_linear}. To generate such non-illusion variants, we consider standard techniques \cite{geier2008straightness,bakshi2020tiny} such as dot-insertion (Fig.~\ref{fig:non_linear} (b)), changing orientation (Fig.~\ref{fig:non_linear} (c), and applying non-linear transformation (Fig.~\ref{fig:non_linear} ( d)). Since there are only limited number of such modifications that make the illusory effects disappear, the number of non-illusion images are comparatively small. Note that a large number of natural images can be considered to be non-illusion images but those are significantly different from illusion images. Our non-illusion images are visually similar to illusion images and thus, identifying illusion images is not a trivial task. The dataset is validated by conducting standard psychophysical experiments supervised by domain experts as described below.

%%%%%%%%%%%%%%%%%%%%%%%%%%%%%%%%%%%%%%%%%%%%%%%%%%%%%%%%%%%%%%%%%%%%%%%%%%%%%%%%%%%%%%%%%%%%%%%%%%
%%%%%%%%%%%%%%%%%%%%%%%%%%%%%%%%%%%%%%%%%%%%%%%%%%%%%%%%%%%%%%%%%%%%%%%%%%%%%%%%%%%%%%%%%%%%%%%%%%

\vspace{-0.2cm}
\subsection{Psychophysical experiment for dataset validation.}
\label{sec:dataset_validation}
\vspace{-0.2cm}

Since the illusory effect is subjective in nature, Psychophysical experiments are designed to measure the illusory effect quantitatively and can also judge the direction of the illusory effect.
We perform psychophysical experiments to validate the dataset using the standard two alternate forced-choice (2AFC) experiment \cite{mitra2018adaptive,shi2013effect}.

%%%%%%%%%%%%%%%%%%%%%%%%%%%%%%%%%%%%%%%%%%%%%%%%%%%%%%%%%%%%%%%%%%%%%%%%%%%%%%%%%%%%%%%%%%%%%%%%%%
\begin{figure*}[!t]
\centering
\includegraphics[scale=0.33]{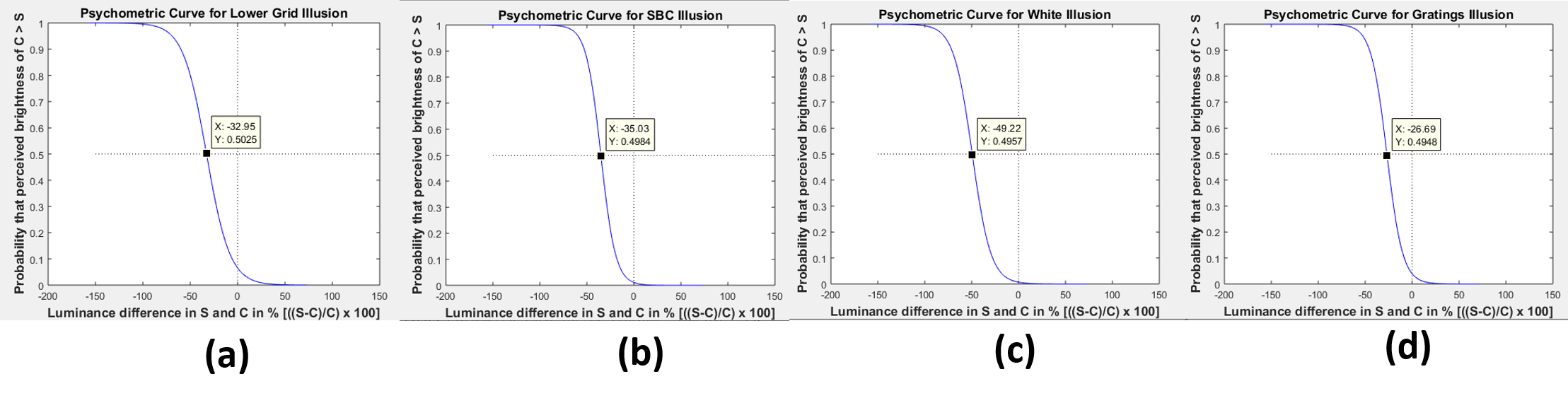}
\vspace{-0.5cm}
\caption{Psychometric curve for (a)Lower grid, (b) SBC, (c) White illusion, (d) Grating illusion taken from the generated dataset.
This indicates the probability that the perception agrees with the reality as a function of the real difference in luminance between the standard (S) and the comparator (C).
The illusory reduction for Grating, White illusion, SBC, Lower grid are 26.69, 49.22, 35.03 and 32.95 respectively. }
\vspace{-0.6cm}
\label{fig:psychometric_curve}
\end{figure*}
%%%%%%%%%%%%%%%%%%%%%%%%%%%%%%%%%%%%%%%%%%%%%%%%%%%%%%%

\textbf{Experimental Setup}
A psychophysical experiment consists of a number of components like stimulus (comparator and standard), task, method, analysis and measure. Two types of procedures exist in psychophysical experiments. In the first type, the comparator is displayed on the computer screen and the observer has to adjust the contrast within a predefined limit to achieve the perceivable intensity as compared with the standard. For this type of procedure, the task and the method are collectively termed as `method of adjustment'.
Nowadays the preferred approach is to display the comparator and standard in a random fashion on a computer screen for a short duration of time. The observer has to decide the relative brightness of the comparator with respect to that of the standard and indicate his/her choice by pressing a key. This is known as 2 Alternate Forced Choice (2AFC) experiment~\cite{mitra2018adaptive}.

\begin{figure}[!t]
   \begin{minipage}{0.48\textwidth}
    \centering
    \includegraphics[scale=0.25]{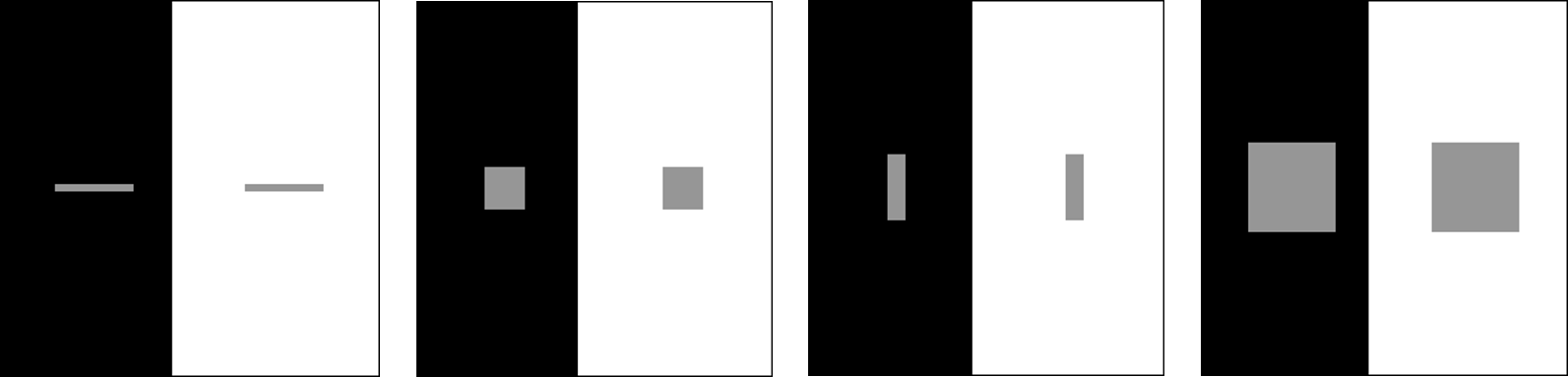}
    \vspace{-0.2cm}
    \caption{\small{Different types of SBC illusions selected as the comparator. The intensity level of the target is always 150. However the length and width of the target varies widely.}}
    %\vspace{-0.2cm}
    \label{fig:exp_setup}
   \end{minipage}\hfill
   \begin{minipage}{0.48\textwidth}
    \centering
    \includegraphics[scale=0.2]{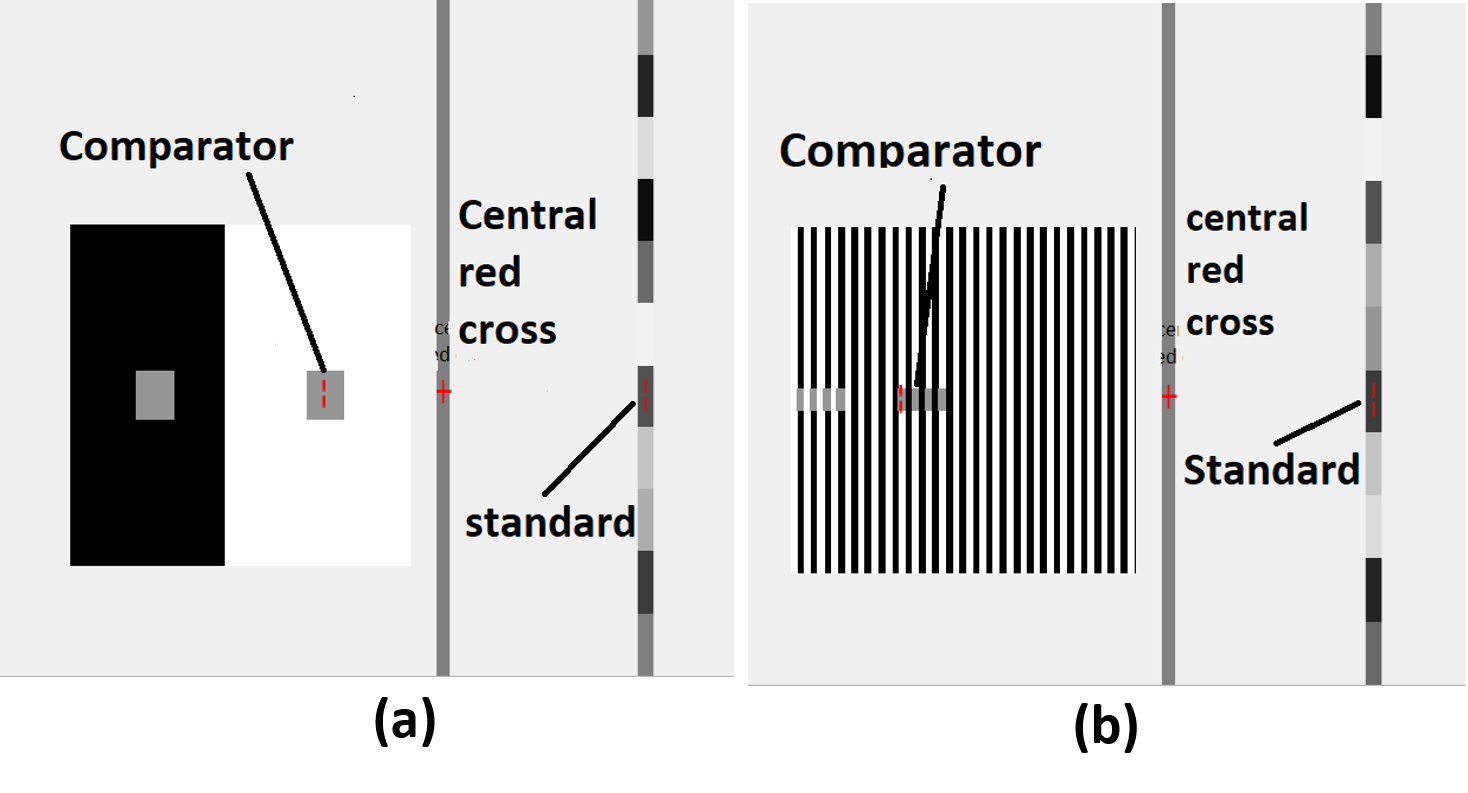}
    \vspace{-0.3cm}
    \caption{\small{Experimental setup for two-alternative forced-choice experiment for (a) SBC  and (b) White illusion. Darker regions (red marked) are compared with the standard randomly multiple times to get the data.}}
    \vspace{-0.4cm}
    \label{fig:2ac_exp}
   \end{minipage}
%\vspace{-0.2cm} 
\end{figure}

In this experiment, seven human subjects, four men and three women (all in age range of 23 to 68 and in perfect health condition) in our case, visually compare the brightness of two target patches: a standard (`S') and a comparator (`C').
We conduct a 2AFC experiment for comparing the brightness between the standard and the comparator in order to measure the illusory enhancement or illusory reduction. The subjects are instructed to provide their feedback by pressing the key marked ONE when the comparator appeared to be brighter than the standard and the key marked as TWO, otherwise. Comparing the `comparator' with the `standard' under various contexts help us identifying the effect of illusion. The comparator (as shown in Fig.~\ref{fig:2ac_exp}) is an intensity-based illusion and it is selected from four types of brightness illusions: SBC, White illusion, gratings, and grid illusions. For the psychophysical experiment, we randomly choose 700 SBC, 400 White illusions, 400 grid illusions, and 400 grating illusions. Since the illusory dots are not physically present in the Hermann grid illusions, they can not be validated using this experiment.

% In our experiments, we consider a segmented vertical band as multiple standards as shown in Fig.~\ref{fig:2ac_exp}. For example, our segmented vertical band is divided into 11 segments of varying intensity values of 13, 36, 59, 82, 105, 128, 150, 173, 196, 219, and 242. These values are kept fixed throughout the experiments. In a single experiment, the same type of illusion e.g., SBC illusion is selected as the comparator as shown in Fig.~\ref{fig:2ac_exp} (a).
% In one set of experiment, the same type of the illusion, say SBC illusion is selected as the comparator. One such set of SBC illusion is shown in Fig.~\ref{fig:exp_setup}. 
% The intensity level of the standard is fixed at 150. However, it appears either lighter or darker depending on the background intensity. In all trials, the darker target is selected as the comparator and its intensity is compared with the standard. In our experiments, the illusory enhancement or decrement are measured at the \textit{point of subjective equality}, i.e., the actual intensities of the comparator and the standard, when these appear to be the same to the subject, from the different sets of psychometric curves drawn for two different illusions as shown in Fig.~\ref{fig:psychometric_curve}. Here, as we always measure the illusory effect of the darker target, we calculate the average illusory decrement for each type of illusion under consideration.

% %%%%%%%%%%%%%%%%%%%%%%%%%%%%%%%%%%%%%%%%%%%%%%%%%%%%%%%%%%%%%%%%%%%%%%%%%%%%%%%%%%%%%%%%%%%%%%%%%%
% \begin{figure*}[!t]
% \centering
% \includegraphics[scale=0.72]{figures/2afc_exp.png}
% \vspace{-0.2cm}
% \caption{\textcolor{blue}{ Illusory  reduction of the gray target patch placed on white background for SBC illusion. The illusory reduction of 7 different subjects is shown from figure (a) to figure (g).}}
% \vspace{-0.4cm}
% \label{fig:2afc_sbc}
% \end{figure*}
%%%%%%%%%%%%%%%%%%%%%%%%%%%%%%%%%%%%%%%%%%%%%%%%%%%%%%%%%%%%%%%%%%%%%%%%%%%%%%%%%%%%%%%%%%%%%%%%%%
\begin{table}[!t]
\caption{ Quantitative Illusory Reduction for different illusions}
\centering
\scalebox{0.55}{
\begin{tabular}{ccccc}
    \hline
    %Illusion & \multirow{Quantitative Illusory Reduction}\\
    % & \multirow{4}{*}{Quantitative Illusory Reduction} \\
    Subject & SBC Illusion & White Illusion & Grating Illusion & Grid Illusion\\
    \hline
    subject 1 & 35.03 & 49.22 & 27.11 & 32.95 \\
    \hline
    subject 2 & 50.47 & 69.24 & 38.79 & 41.71 \\
    \hline
    subject 3 & 44.21 & 62.15 & 47.97 & 49.63 \\
    \hline
    subject 4 & 47.55 & 58.39 & 33.37 & 46.3 \\
    \hline
    subject 5 & 45.46 & 49.63 & 37.12 & 31.28 \\
    \hline
    subject 6 & 32.95 & 42.54 & 26.69 & 29.61 \\
    \hline
    subject 7 & 45.88 & 52.14 & 30.45 & 30.45 \\
    \hline
    Avg. illusory effect & 43.08 & 54.76 & 34.50 & 37.42 \\
    \hline
\end{tabular}}
\vspace{-0.6cm}
\label{Tab: quantitative_illusory_reduction}
\end{table}

\textbf{Psychometric Results}
The psychometric curves (using seven observers) for SBC, White illusion, gratings illusion and lower grid illusion are provided in Fig.~\ref{fig:psychometric_curve}.
% In Fig.~\ref{fig:2afc_sbc} we have shown the psychophysical test for SBC illusion with seven observers. 
We have also done similar psychophysical experiments for White, grating and grid illusion and the illsuory reduction is shown in Table.~\ref{Tab: quantitative_illusory_reduction}, which shows similar trend holds for other illusions and we observe similar psychometric curve for White, grating and grid illusion as well.
Illusory decrement indicates the quantitative measure of perceived darkness. For example, given the standard grayscale value to be 150, illusory reduction of 32.95 indicates that the target is perceived to be grayscale value of 117.05 and looks darker. The illusory reduction of the same intensity comparator of grayscale value 150 for White illusion is maximum (49.22), for gratings illusion is minimum (27.11), and it lies within this maximum and minimum limit for SBC (35.03) and Lower-grid illusion (32.95) as shown in Fig.~\ref{fig:psychometric_curve}. These psycophysical experiments statistically validate the dataset. A detailed description of the experimental setup is provided in the supplementary material.

% Fig.~\ref{fig:psychometric_curve} shows a sample psychometric curves for the SBC and White's illusions. Illusory reduction indicates the quantitative measure of perceived darkness. For example, given the standard gray intensity value of 150, an illusory reduction of 35 indicates that the standard is perceived as an intensity value of 185 and looks darker. A detailed description of the experimental setup is provided in the supplementary material.

% The illusory reduction of the same intensity comparator (fixed at grayscale value 150) for white illusion is maximum (49.22), for gratings illusion is minimum (27.11), and it lies within this maximum and minimum limit for SBC (35.03) and Lower-grid illusion (32.95) as shown in Fig.~\ref{fig:psychometric_curve}. A detailed description of the experimental setup is provided in the supplementary material.

\vspace{-0.4cm}
\section{Tasks: Illusion Identification, Localization and Generation}
\label{sec:methodology}
\vspace{-0.4cm}

% \subsection{Illusion identification}
%\label{sec:illusion_det}
\textbf{1. Illusion identification.}
Illusion identification is posed as a binary decision problem to identify whether an image is a brightness illusion or not. Although we have a relatively large collection of illusion images, non-illusion images are relatively scarce. To combat the nuisance of data imbalance, we have performed data augmentation since any linear transformation on a non-illusion image tends to be a non-illusion itself. Recall that we only consider carefully created non-illusion images which are visually similar to illusion images instead of widely available natural images (Fig.~\ref{fig:non_linear}). For classification, we consider three widely explored variants of CNNs: 1) AlexNet~\cite{krizhevsky2012imagenet} which is a five-layer CNN, 2) Resnet-18~\cite{he2016deep} which is an 18-layer CNN with residual connections between the layers, and 3) SqueezeNet~\cite{iandola2016squeezenet} which is designed for low-capacity devices with very small number of parameters. For training, we use binary cross-entropy loss and stochastic gradient descent optimization with a learning rate of 0.001 and a momentum of 0.9. Details of the networks are provided in the supplemental material.

% \vspace{-0.2cm}
%\subsection{Illusion Localization}
\textbf{2. Illusion Localization.}
\label{sec:illusion_loc} 
We cast illusion localization as a binary segmentation problem, where we localize the region corresponding to perceived darkness in the image. %(Fig.~\ref{fig:methodology})
Recall that we follow the convention to localize the perceived \textit{darkness} and the complementary lightness regions can be inferred by matching the intensity value. We consider the UNet~\cite{ronneberger2015u} for segmentation. The network contains two paths. The first path is the contraction path (as an encoder) which captures the high-level details such as the larger context of the image. The second path is the expanding path (as a decoder) which captures the low-level details such as edges and local textures. This network is suitable for illusion localization as it can capture local brightness variation and observe the local brightness regions in a larger context to perceive the illusion.

For training the segmentation network, along with the commonly used mean square error (MSE) loss, we also consider the structural similarity index metric (SSIM) to capture perceptual similarity. Bovik et al.~\cite{wang2004image} introduced the SSIM inspired by the fact that human perception is highly adapted to perceive structured information in a scene. We consider a differentiable version of SSIM which is shown to be effective for several low-level tasks including deblurring and denoising ~\cite{zhao2016loss}. We consider a convex combination of MSE loss ($L_{MSE}$ and SSIM loss ($L_{SSIM}$)) as the loss function, $L = \alpha L_{MSE} + \beta L_{SSIM}$, where $L_{MSE}$ is mean squared error loss and $L_{SSIM}$~\cite{zhao2016loss} is the SSIM loss. We experimentally set the parameters, $\alpha = 0.4$, $\beta = 0.6$, which produced the best performance (Tab.~\ref{tab:loc_table}). We consider the Adam optimizer with a learning rate of 0.001 and a batch size of 32 images.

\begin{figure}[!t]
\centering
\includegraphics[scale=0.33]{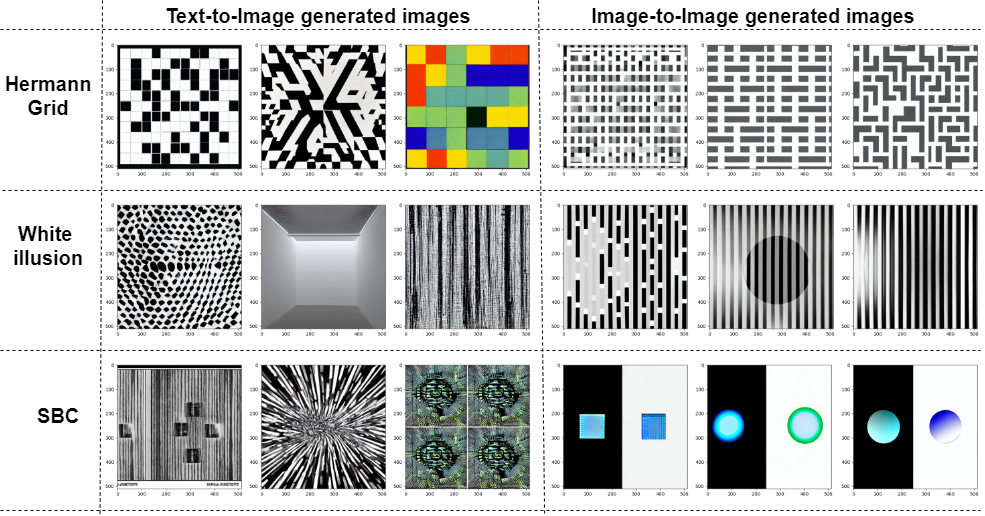}
\vspace{-0.5cm}
\caption{Diffusion model generated illusion images.}
\vspace{-0.5cm}
\label{fig:ill_diffusion}
\end{figure}

\textbf{3. Illusion generation using diffusion model.}
Motivated by the recent success of generative models, we generate illusions using text-to-image~\cite{rombach2022high} and image-to-image diffusion model~\cite{meng2021sdedit}.
To the best of our knowledge, we are first to investigate the ability of generating illusions through diffusion models.
We use stable diffusion~\cite{rombach2022high} for both the cases, where text-to-image diffusion model uses ``class name'' (e.g, Hermann grid illusion, White illusion etc.) as text prompt and image-to-image diffusion model uses both illusion images and text prompt (i.e., class names) as input to generate diverse illusions. 
Since the diffusion models are not trained on illusion dataset (which practically does not exists earlier), it generates somewhat misleading images (Fig.~\ref{fig:ill_diffusion} ). However, image-to-image diffusion model modifies input illusion images guided by text prompts (i.e., class names), thus generates interesting variants of illusions as shown in Fig.~\ref{fig:ill_diffusion}. 
We have tested the generated illusion images using experts and it seems the image-to-image diffusion model generates better illusions than text-to-image diffusion model.
In particular, the essence of the brightness illusions are captured properly for SBC, White illusions (as shown in Fig.~\ref{fig:ill_diffusion}) and qualitatively the rank of quality of generated illusions are as follows: SBC $>$ White illusion $>$ Hermann grid (through observation of experts).
To quantify the generative quality of these images, we train a classifier (2 layer MLP on top of ResNet feature extractor) using the text-to-image and image-to-image generated images for each of the five class of illusion and test on the same set of actual illusion images. The classification performance is lower for classifier trained on just text-to-image generated images, but significantly improves when trained on image-to-image generated images as shown in Tab.~\ref{tab:diff_compare}. 

\vspace{-0.4cm}
\section{Experiments}
\vspace{-0.4cm}
We perform the following experiments to evaluate various aspects of our approach: 1) illusion identification, 2) illusion localization, 3) generalization of illusion localization to transition between illusions, and 4) analysis of network layers for the models trained on natural images vs. illusion images. 
Our dataset and code can be downloaded from the public domain \href{https://github.com/aniket004/BRI3L}{https://github.com/aniket004/BRI3L}. 
% A simple evaluation of the methods using the dataset can be evaluated using the google colab notebook: \url{https://colab.research.google.com/drive/1g4Ov5Cbx4nIzd-QxabmtuFC9A-rMdrO0#scrollTo=WUlmxxDmtHqd}

\begin{table*}[]
\begin{center}
    \begin{minipage}{.33\linewidth}
    \centering
    \caption{\small Test accuracy of illusion identification.}
    \label{tab:det_table}    
    \scalebox{0.6}{
    \begin{tabular}{ccccc}
    \hline
    \textbf{Network} & \textbf{Accuracy} & \textbf{F1-score} & \textbf{Precision} & \textbf{Recall} \\ \hline
    \textbf{ResNet-18} & 0.9956            & 0.9948            & 1.00               & 0.9897          \\ \hline
    \textbf{AlexNet} & 0.9973 & 0.9969 & 0.9938 & 1.00 \\ \hline
    \textbf{SqueezeNet} & 0.9956 & 0.9948 & 1.00 & 0.9897 \\ \hline
    \end{tabular}}
    \end{minipage}
    \hfill
    \begin{minipage}{.33\linewidth}
    \centering
    \caption[\small]{Illusion classification using diffusion model generated images. 
    % Generated images using Image-to-image diffusion model only (I2I only) performs better compared to text-to-image (T2I only) and combined I2I and T2I. The classification accuracy is reported for 5-way classification problem and same test set comprising of actual illusion images.
    }
    \scalebox{0.7}{
    \begin{tabular}{cc}
    \hline
    \textbf{Method} & \textbf{Accuracy}\\ \hline
    \textbf{T2I only}& {$37.85\%$} \\ \hline
    \textbf{I2I only} & {$87.67\%$} \\ \hline
    \textbf{T2I+I2I} & {$81.32\%$ } \\ \hline
    \end{tabular}}
    \label{tab:diff_compare}
    \end{minipage}
    \hfill    
    \begin{minipage}{.33\linewidth}
    \centering
    \caption{\small Test accuracy for illusion classification.}
    \label{tab:clf_table}
     \scalebox{0.7}{
    \begin{tabular}{ccccc}
    \hline
    \textbf{Network} & \textbf{Accuracy} & \textbf{F1-score} & \textbf{Precision} & \textbf{Recall} \\ \hline
    \textbf{ResNet-18} & 0.9568           & 0.9964          & 0.9995               & 0.9934          \\ \hline
    \textbf{AlexNet} & 0.8952 & 0.9455 & 1.00 & 0.8967 \\ \hline
    \textbf{SqueezeNet} & 0.9310 & 0.9738 & 1.00 & 0.9490 \\ \hline
    \end{tabular}}
    \end{minipage}
\end{center}
\vspace{-0.5cm}
\end{table*}

\textbf{Metrics.} 
% Due to the lack of large-scale datasets for illusion understanding, we demonstrate the effectiveness of our approach for illusion identification and localization on the proposed brightness illusion dataset. 
% The dataset contains 22366 illusion images of size 256$\times$256 consisting of 4160 SBC illusions, 637 White illusions, 1024 Hermann grid illusions, 6350 induced grating illusions, and 10195 upper and lower grid illusions. There are 1149 non-illusion images. Each illusion image is annotated with a binary mask corresponding to the illusory regions. 
For illusion identification, we consider \textit{Accuracy}, \textit{F1-score}, \textit{Precision}, \textit{Recall} as metrics and for illusion localization, we consider pixel-wise \textit{F1-score} and \textit{mean Intersection Over Union (mIOU) metric} which is commonly used to evaluate segmentation approaches \cite{ronneberger2015u}. We provide the description of the dataset creation in the supplementary material.

\textbf{1. Illusion identification.} We have 22,366 illusion images and 1149 non-illusion images in the dataset. We augment the non-illusion images to create a balanced set for training. We consider common data-augmentation approaches such as random resized cropping, center-cropping, vertical flip, horizontal flip to make around 3000 non-illusions, from which 2000 used for training and 1000 for testing. Then we randomly sample 1000 illusions from all variants of illusions for training and the rest of the images are used for testing. We consider three standard models - AlexNet, ResNet-18, and SqueezeNet. We achieve 99.56\% accuracy for illusion identification as shown in Table~\ref{tab:det_table}. We also perform illusion classification to correctly classify illusion images into one of the five classes (White, Hermann grid, grid illusion, SBC, and grating illusion). We consider 500 images from each class for training and test on the remaining images. The results are shown in Tab.~\ref{tab:clf_table}.
% \begin{figure}[!htb]
%   \begin{minipage}{0.48\textwidth}
%      \centering
%      \includegraphics[width=.7\linewidth]{example-image-a}
%      \caption{Interpolation for Data 1}\label{Fig:Data1}
%   \end{minipage}\hfill
%   \begin{minipage}{0.48\textwidth}
%      \centering
%      \includegraphics[width=.7\linewidth]{example-image-b}
%      \caption{Interpolation for Data 2}\label{Fig:Data2}
%   \end{minipage}
% \end{figure}
%%

% \begin{table}
% \scalebox{0.7}{
% \begin{tabular}{|c|c|c|c|c|}
% \hline
% \textbf{Network} & \textbf{Accuracy} & \textbf{F1-score} & \textbf{Precision} & \textbf{Recall} \\ \hline
% \textbf{ResNet-18} & 0.9956            & 0.9948            & 1.00               & 0.9897          \\ \hline
% \textbf{AlexNet} & 0.9973 & 0.9969 & 0.9938 & 1.00 \\ \hline
% \textbf{SqueezeNet} & 0.9956 & 0.9948 & 1.00 & 0.9897 \\ \hline
% \end{tabular}}
% \caption{\small Test accuracy of illusion identification.}
% \label{tab:det_table}
% \end{table}

% %% classification table
% \begin{table}[!t]
% \centering
% \scalebox{0.7}{
% \begin{tabular}{|c|c|c|c|c|}
% \hline
% \textbf{Network} & \textbf{Accuracy} & \textbf{F1-score} & \textbf{Precision} & \textbf{Recall} \\ \hline
% \textbf{ResNet-18} & 0.9568           & 0.9964          & 0.9995               & 0.9934          \\ \hline
% \textbf{AlexNet} & 0.8952 & 0.9455 & 1.00 & 0.8967 \\ \hline
% \textbf{SqueezeNet} & 0.9310 & 0.9738 & 1.00 & 0.9490 \\ \hline
% \end{tabular}}
% \caption{\small Test accuracy for illusion classification.}
% \label{tab:clf_table}
% \end{table}

%%
\begin{table}[!t]
\begin{center}
    \begin{minipage}{1.0\linewidth}
    \centering
    \caption[\small]{Test accuracy of illusion localization with various combinations of $\alpha$ and $\beta$. Recall the loss function is $L = \alpha L_{MSE} + \beta L_{SSIM}$.}
    \label{tab:loc_table}   
    \scalebox{0.7}{
    \begin{tabular}{ccccc}
    \hline
    \textbf{$\alpha$} & \textbf{$\beta$} &  \textbf{F1-score} &  \textbf{mIoU} \\ \hline
    1                 & 0         & 0.5666    & 0.61          \\ \hline
    0                 & 1      & 0.3167    & 0.46          \\ \hline
    0.5               & 0.5           & 0.4386    & 0.5           \\ \hline
    0.6               & 0.4       & 0.4395    & 0.53          \\ \hline
    0.4               & 0.6      & \textbf{0.7689}   & \textbf{0.75} \\ \hline
    \end{tabular}}
    \end{minipage}
    % \hfill
    % \begin{minipage}{.45\linewidth}
    % \centering
    % \caption[\small]{Illusion classification using diffusion model generated images. 
    % % Generated images using Image-to-image diffusion model only (I2I only) performs better compared to text-to-image (T2I only) and combined I2I and T2I. The classification accuracy is reported for 5-way classification problem and same test set comprising of actual illusion images.
    % }
    % \scalebox{0.7}{
    % \begin{tabular}{cc}
    % \hline
    % \textbf{Method} & \textbf{Accuracy}\\ \hline
    % \textbf{T2I only}& {$37.85\%$} \\ \hline
    % \textbf{I2I only} & {$87.67\%$} \\ \hline
    % \textbf{T2I+I2I} & {$81.32\%$ } \\ \hline
    % \end{tabular}}
    % \label{tab:diff_compare}
    % % \caption{ \small Transferring illusions across other brightness illusions: testing on particular illusion as mentioned in the rows (from our dataset) while training the segmentation model for dark region localization on all other types except that particular illusion.}
    % % \label{tab:ill_transfer}
    % % \scalebox{0.6}{
    % % \begin{tabular}{ccc}
    % % \hline
    % % \textbf{Illusion} & \textbf{Accuracy} & \textbf{mIoU} \\ \hline
    % % Hermann grid      & 0.6905            & 0.43          \\ \hline
    % % Induced grating   & 0.9206            & 0.49          \\ \hline
    % % Lower grid        & 0.7991            & 0.43          \\ \hline
    % % Upper grid        & 0.8546            & 0.44          \\ \hline
    % % SBC               & 0.8805            & 0.48          \\ \hline
    % % White             & 0.9745            & 0.5           \\ \hline
    % % \end{tabular}}
    % \end{minipage}
\end{center}
\vspace{-0.6cm}
\end{table}

\textbf{2. Illusion localization.} We consider the UNet ~\cite{ronneberger2015u} for illusion localization.  We consider 13419 images for training and test on remaining 8947 images. A convex combination of MSE and SSIM loss is used to train the network. We consider various values of the parameters $\alpha$ and $\beta$ and observe that the best segmentation performance is achieved when $\alpha = 0.4$ and $\beta=0.6$, i.e., $L = 0.4 L_{MSE} + 0.6 L_{SSIM}$. The results are shown in Table.~\ref{tab:loc_table} where we achieve 84.37\% accuracy and a mIoU score of 0.75. Note that $\alpha$ = 1, $\beta$ = 0 implies considering the MSE loss, i.e.,  $L = L_{MSE}$ and $\alpha$ = 0, $\beta$ = 1 implies only the SSIM loss, i.e., $L = L_{SSIM}$. The results justify the importance of perceptual cues in the form of SSIM loss for illusion localization.

\textbf{3. Transition from assimilation illusion to contrast illusion.} Here we test the effectiveness of illusion localization while an illusion transition from assimilation to contrast. Note that the model has never seen such transitions. We consider two popular illusion transitions: Howe stimulus~\cite{howe2001comment} and shifted White to checkerboard transition~\cite{white1979new}.
% These type of sequences of images where transition from contrast to assimilation (or vice versa) happens seem to be infer the essence of unified modelling of contrast and assimilation to better mimic the visual perception~\cite{robinson2007explaining, gilchrist1999anchoring}.

\begin{figure*}[]
\centering
\includegraphics[scale=0.3]{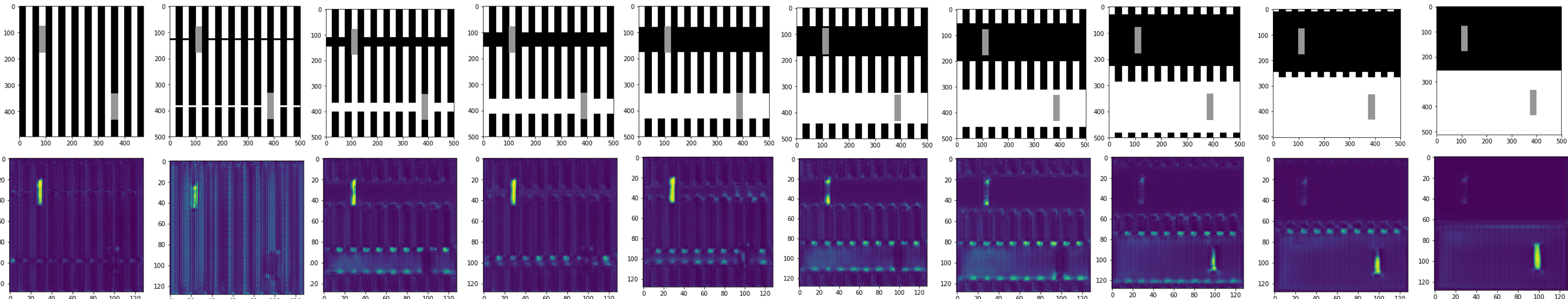}
\vspace{-0.2cm}
\caption{Howe's Illusion: Transition from White illusion to SBC. For the images in the upper row in particular, the model localizes the upper left test patch as the dark one, which akin to human perception~\cite{howe2001comment} vacillates and finally localizes in the lower right test patch as the dark one.}
\vspace{-0.4cm}
\label{fig:howe}
\end{figure*}

\textbf{Howe illusion.} Howe illusion~\cite{howe2001comment} is a transition from White illusion to SBC as shown in Fig.~\ref{fig:howe}, where there is a shift of perceived darkness by changing the background. Initially, the leftmost figure in the top row of Fig.~\ref{fig:howe} is the White illusion. However, as the width of both the dark and white line crossing the patches increases along the row from left to right, the darker patch gets brighter and finally turns into an SBC. The bottom row shows the localization output which is consistent with human observations. Therefore, our learned model can identify such transition in perceived darkness akin to human perception as shown in Fig.~\ref{fig:howe}.

%%%%%%%%%%%%%%%%%%%%%%%%%%%%%%%%%%%%%%%%%%%%%%%%%%%%%%%%%%%%%%%%%%%%%%%%%%%%%%%%%%%%%%%%%%%%%%%%%%
%%%%%%%%%%%%%%%%%%%%%%%%%%%%%%%%%%%%%%%%%%%%%%%%%%%%%%%%%%%%%%%%%%%%%%%%%%%%%%%%%%%%%%%%%%%%%%%%%%

\begin{figure}[t]
   \begin{minipage}{0.48\textwidth}
    % \centering
    % \includegraphics[scale=0.22]{figures/crossover_shifted_white.png}
    % \caption{Psychophysical experiment validating the transition of illusory effect in shifted White illusion with varying aspect ratio as predicted by the model in Fig.~\ref{fig:shifted_white}.}
    % \label{fig:psyco_shifted_white}
    \centering
    \includegraphics[scale=0.2]{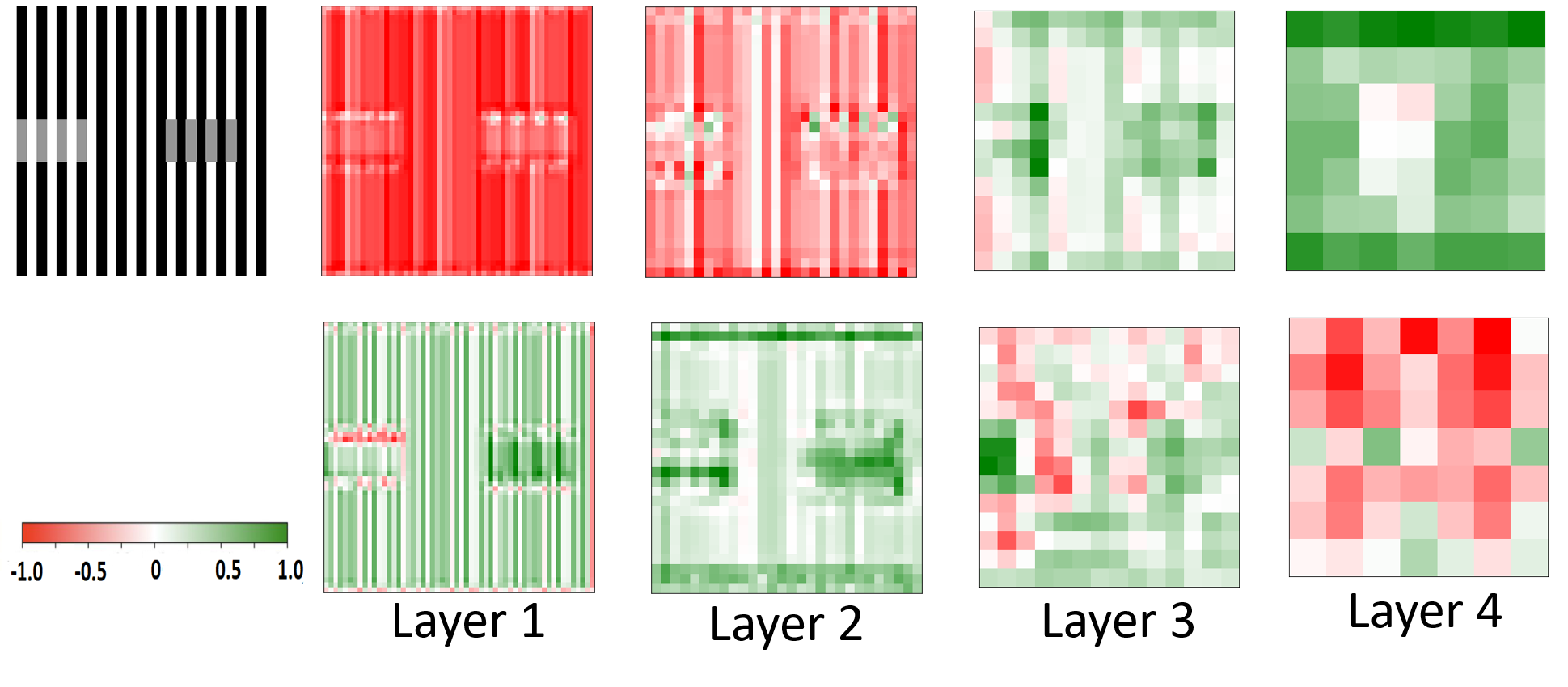}
    \vspace{-0.3cm}
    \caption[\small]{Gradcam attributes of the ResNet18 layers for the White illusion. Upper row: model trained with natural images using the ImageNet dataset. Lower row: model trained with illusions.}
    \label{fig:white_cam}
   \end{minipage}\hfill
   \begin{minipage}{0.48\textwidth}
    \centering
    \includegraphics[scale=0.18]{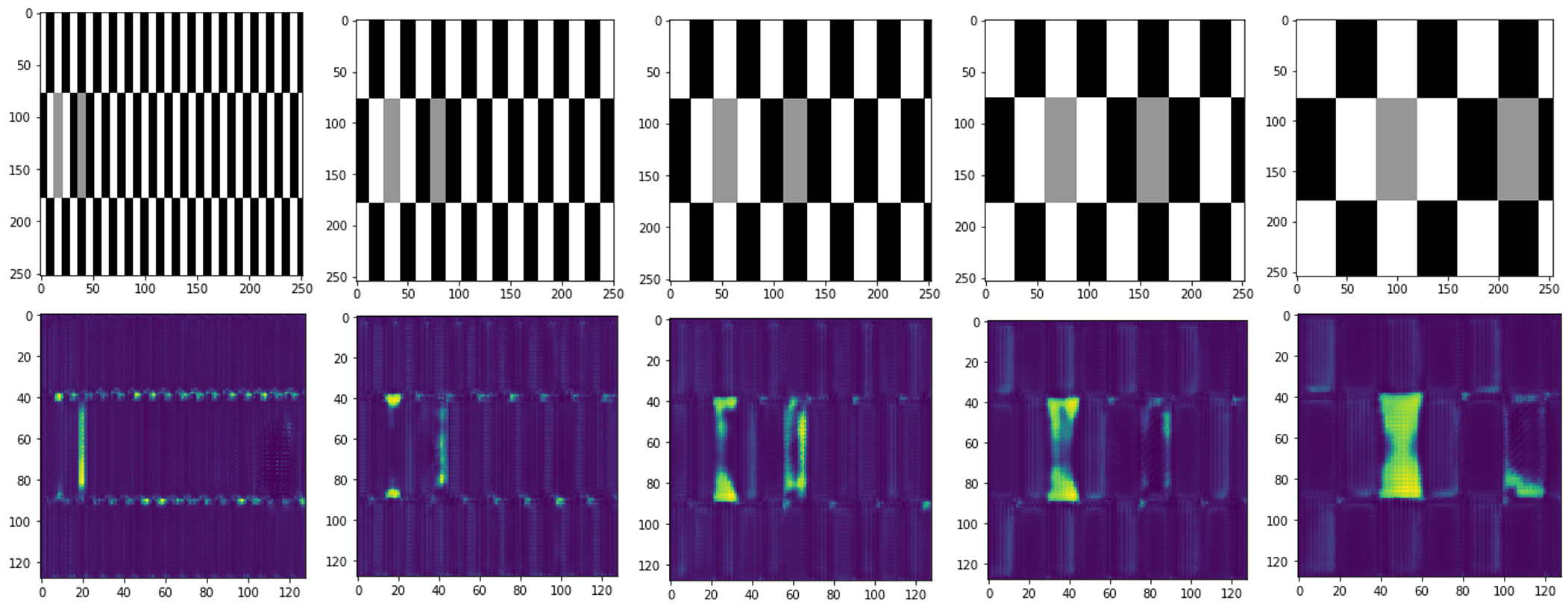}
    \vspace{-0.3cm}
    \caption[\small]{Transition from assimilation to contrast in Shifted White stimulus: Inversion in perceived darkness is observed by the learned model while reducing the aspect ratio (rightmost image). Images (upper row) and their corresponding response (lower row) by the learned model.}
    \label{fig:shifted_white}
   \end{minipage}
\vspace{-0.6cm}
\end{figure}

%\vspace{-0.4cm}
\textbf{Brightness Transition in Shifted White illusion with Aspect Ratio Variation.}
\label{sec:transition}
Fig.~\ref{fig:shifted_white} illustrates the shifted White illusion, which is a variant of White illusion~\cite{white1979new}. An intriguing feature of shifted White illusion, we found, is the inversion in the perceived brightness/darkness of the patch when we decrease the aspect ratio of the strips as shown in Fig.~\ref{fig:shifted_white}. In the leftmost figure in the top row (Fig.~\ref{fig:shifted_white}) the right patch appears darker. However, if the aspect ratio of the patch is reduced along the row, i.e. as the stimulus shifts from Shifted White to a Checkerboard, then in the rightmost figure the darker patch turns out to be lighter as shown in Fig.~\ref{fig:shifted_white}. Our learned model adequately mimics such inversion of perceived darkness as shown in Fig.~\ref{fig:shifted_white}.

% A psychometric curve for this transition using the same protocol followed in Sec.~\ref{sec:dataset_validation} has been shown in Fig.~\ref{fig:psyco_shifted_white}, which verifies our finding by the 2AFC experiment.

% \textbf{5. Complex brightness illusions.} Our learned model can localize the perceived darkness in more complex brightness illusions which are created by combining the contrast and assimilation illusions as shown in Fig.~\ref{fig:success_test_case}. Note that these illusions exhibit more than three intensity levels which are different from the training images with a maximum of three intensity values. 
% %%%%%%%%%%%%%%%%%%%%%%%%%%%%%%%%%%%%%%%%%%%%%%%%%%%%%%%%%%%%%%%%%%%%%%%%%%%%%%%%%%%%%%%%%%%%%%%%%%
% \begin{figure}[!t]
% \centering
% \vspace{-0.2cm}
% \includegraphics[scale=0.22]{figures/nat_img_ill.png}
% \vspace{-0.2cm}
% \caption{\textcolor{blue}{Model response (lower row) on natural images (upper row): (a), (b), (c) natural images showing Mach band, (d) Illusory regions identified in illusions generated from natural image by ~\cite{hirsch2020color}.}}
% \vspace{-0.2cm}
% \label{fig:nat_img_ill}
% \end{figure}
% %%%%%%%%%%%%%%%%%%%%%%%%%%%%%%%%%%%%%%%%%%%%%%%%%%%%%%%%%%%%%%%%%%%%%%%%%%%%%%%%%%%%%%%%%%%%%%%%%%

\textbf{4. Network analysis.} We investigate the feature maps learned by CNN models while identifying the illusions. We compute layer-wise attribution using GradCam~\cite{selvaraju2017grad} to analyze what features are encoded across the CNN layers. This approach, for a given target output, computes the contribution of the image regions for the final prediction across the layers. We use the Captum library~\cite{kokhlikyan2020captum} for computing the layer-wise attributions. We compare the layer-wise attributions of the ResNet model trained on natural images and trained on illusion images. The result for White illusion is shown in Fig.~\ref{fig:white_cam}. Note that there is a significant difference between the contribution of the image regions across the layers between the network trained with natural images (top row) and illusions (bottom row). We also notice that the lower layers tend to focus on low-level illusory features such as the boundary between two regions with various intensities. The top layers learn more abstract features. The differences between models trained with natural images and illusions are more prominent in the lower layers. These observations indicate that perception of brightness illusions is primarily a low and mid-level vision task.

% %%%%%%%%%%%%%%%%%%%%%%%%%%%%%%%%%%%%%%%%%%%%%%%%%%%%%%%%%%%%%%%%%%%%%%%%%%%%%%%%%%%%%%%%%%%%%%%%%%
% \begin{figure}[!t]
% \centering
% %\vspace{-0.2cm}
% \includegraphics[scale=0.265]{LaTeX/figures/grid_gradcam.png}
% \vspace{-0.2cm}
% \caption[\small]{Gradcam attributes on Hermann grid image in layers of ResNet18. Lower row: attributes with finetuned model. Upper row: attributes with pretrained model.}
% \vspace{-0.3cm}
% \label{fig:grid_cam}
% \end{figure}
% %%%%%%%%%%%%%%%%%%%%%%%%%%%%%%%%%%%%%%%%%%%%%%%%%%%%%%%%%%%%%%%%%%%%%%%%%%%%%%%%%%%%%%%%%%%%%%%%%%

% \begin{figure}[!t]
%    \begin{minipage}{0.48\textwidth}
%     \centering
%     \includegraphics[scale=0.2]{figures/white_gradcam.png}
%     \vspace{-0.2cm}
%     \caption[\small]{Gradcam attributes of the ResNet18 layers for the White illusion. Upper row: model trained with natural images using the ImageNet dataset. Lower row: model trained with illusions.}
%     \vspace{-0.3cm}
%     \label{fig:white_cam}
%    \end{minipage}\hfill
%    \begin{minipage}{0.48\textwidth}
%     \centering
%     \includegraphics[scale=0.2]{figures/grating_cam.png}
%     \vspace{-0.2cm}
%     \caption[\small]{Gradcam attributes of the ResNet18 layers for the grating illusion. Upper row: model trained with natural images using the ImageNet dataset. Lower row: model trained with illusions.}
%     \vspace{-0.3cm}
%     \label{fig:grating_cam}
%    \end{minipage}
% \end{figure}

\vspace{-0.4cm}
\section{Conclusion}
\vspace{-0.4cm}

Understanding and evaluating visual illusions is a challenging task in computational neuroscience. 
% Existing methods generally use deterministic filtering based approach to model visual perception due to lack of benchmark dataset.
To facilitate exploration in illusion understanding, we have introduced a large-scale dataset of 22,366 images containing five types of brightness illusions. We have annotated each image with the binary mask corresponding to illusory regions in the image. We have validated the dataset by conducting standard psychophysical experiments involving human experts.
Next, we benchmark the dataset with deep learning based data-driven approach for illusion identification and localization for brightness illusions.
% The proposed approach has achieved 99\% accuracy on illusion identification and 75\% mIoU accuracy on illusion localization. 
We have conducted extensive experiments evaluating various aspects of our approach. 
% Our experiments show that the proposed approach can generalize to various types of novel brightness perceptions.

% \section{REFERENCES}
% \label{sec:refs}

% List and number all bibliographical references at the end of the
% paper. The references can be numbered in alphabetic order or in
% order of appearance in the document. When referring to them in
% the text, type the corresponding reference number in square
% brackets as shown at the end of this sentence \cite{C2}. An
% additional final page (the fifth page, in most cases) is
% allowed, but must contain only references to the prior
% literature.

% References should be produced using the bibtex program from suitable
% BiBTeX files (here: strings, refs, manuals). The IEEEbib.bst bibliography
% style file from IEEE produces unsorted bibliography list.
% -------------------------------------------------------------------------
%\scriptsize
\bibliographystyle{IEEEbib}
\setlength\itemsep{0em}
\bibliography{name}
%\bibliography{strings,refs}

% Template for ICIP-2024 paper; to be used with:
%          spconf.sty  - ICASSP/ICIP LaTeX style file, and
%          IEEEbib.bst - IEEE bibliography style file.
% --------------------------------------------------------------------------
% \documentclass{article}
% \usepackage{spconf,amsmath,graphicx}
% \usepackage{url}
% \usepackage{booktabs} 
% \usepackage{times}
% \usepackage{mathptmx}
% \usepackage[T1]{fontenc}
% \usepackage[utf8]{inputenc}
% \usepackage{hyperref}

% Example definitions.
% --------------------
\def\x{{\mathbf x}}
\def\L{{\cal L}}

% Title.
% ------
\title{BRI3L: A brightness illusion image dataset for identification and localization of regions of illusory perception}
%
% Single address.
% ---------------
% \name{Author(s) Name(s)\thanks{Thanks to XYZ agency for funding.}}
% \address{Author Affiliation(s)}

% \name{Aniket Roy$^{1}$, Anirban Roy$^{2}$, Soma Mitra$^{3}$, Kuntal Ghosh$^{4}$ }
% \address{$^{1}$Johns Hopkins University, $^{2}$SRI International, $^{3}$CDAC Kolkata, $^{4}$Indian Statistical Institute}

% \author{%
%   Aniket Roy\\
%   Department of Computer Science\\
%   Johns Hopkins University\\
%   Baltimore, USA \\
%   \texttt{ank.roy4@gmail.com} \\
%    \And
%    Anirban Roy \\
%    SRI International \\
%    USA \\
%    \texttt{anirbanroy.ti@gmail.com } \\
%    \AND
%    Soma Mitra \\
%    CDAC, Kolkata \\
%    Kolkata, India \\
%    \texttt{soma.mitra@cdac.in} \\
%    \And
%    Kuntal Ghosh \\
%    Indian Statistical Institute \\
%    Kolkata, India \\
%    \texttt{kuntalghos@gmail.com} \\
% }

%
% For example:
% ------------
%\address{School\\
%	Department\\
%	Address}
%
% Two addresses (uncomment and modify for two-address case).
% ----------------------------------------------------------
% \twoauthors
%  {A. Author-one, B. Author-two\sthanks{Thanks to XYZ agency for funding.}}
% 	{School A-B\\
% 	Department A-B\\
% 	Address A-B}
%  {C. Author-three, D. Author-four\sthanks{The fourth author performed the work
% 	while at ...}}
% 	{School C-D\\
% 	Department C-D\\
% 	Address C-D}

%\begin{document}
%\ninept
%
\maketitle

\section{Supplementary Material}

In the supplementary material, we are providing the following details.
\begin{enumerate}
    \item Dataset curation details 
    \item Details of psychophysical experiments and results.
    \item Details of dataset generation.
    \item Networks details for illusion detection and localization.
    \item More qualitative experimental results.
    \item Generalization to unseen illusions.
    \item Illusion vs natural images.
    \item Test on illusions in natural images.
    \item Network analysis.
\end{enumerate}

\section{Dataset curation details}

Our dataset and code can be downloaded from the public domain \href{https://github.com/aniket004/BRI3L}{https://github.com/aniket004/BRI3L}. More details of dataset maintenance, download and usage have been provided in this website.
A simple evaluation of the methods using the dataset can be evaluated using the google colab notebook: 
\href{https://colab.research.google.com/drive/1g4Ov5Cbx4nIzd-QxabmtuFC9A-rMdrO0#scrollTo=WUlmxxDmtHqd}{notebook}

\section{Psychophysical Experiment : Detail experimental setup, procedure and result.}

\subsection{Background}

Gustav Theodor Fechner~\cite{fechner18601966} was the pioneer to set  the principles and procedures of psychophysics. It is a sub-discipline  of psychology and deals with the relationship between physical stimuli and their subjective correlates or percepts.
In different intensity based visual illusions, the lightness of the test patch is modified which is termed as illusory effect. This illusory effect is subjective in nature. Performing comparison of the intensity of the test patch with a standard patch of known intensity, one can only guess whether the test patch is lighter or darker as compared with the standard. Psychophysical experiments are designed to measure the illusory effect quantitatively and can also judge the direction of the illusory effect.

A psychophysical experiment consists of a number of components like stimulus (comparator and standard), task, method, analysis and measure. Two types of procedures exist in psychophysical experiments. In the first type, the comparator is displayed on the computer screen and the observer has to adjust the contrast within a predefined limit to achieve the perceivable intensity as compared with the standard. For this type of procedure, the task and the method are collectively termed as `method of adjustment'.
Nowadays the preferred approach is to display the comparator and standard in a random fashion on a computer screen for a short duration of time. The observer has to decide the relative brightness of the comparator with respect to that of the standard and indicate his/her choice by pressing a key. This is known as 2 Alternate Forced Choice (2AFC) experiment~\cite{mitra2018adaptive}.

Four types of intensity based illusions namely, Simultaneous Brightness Contrast(SBC), White Illusion (WI), Gratings and Lower Grid illusions with variable parameters ( say length, width, intensity background etc.) are generated randomly and the darker patch is selected as the target patch or comparator.
The two alternate forced choice (2AFC) experimental setup has been adopted to quantify the illusory enhancement or reduction of the selected target patch. The psychometric curve is drawn in Fig. 4 (in the paper) and Fig.~\ref{fig:psychometric_curve_supp} for a specific illusion and the illusory effect of the test patch is quantified at the point of subjective equality (PSE).

%%%%%%%%%%%%%%%%%%%%%%%%%%%%%%%%%%%%%%%%%%%%%%%%%%%%%%%%%%%%%%%%%%%%%%%%%%%%%%%%%%%%%%%%%%%%%%%%%%
\begin{figure}[!t]
\centering
\includegraphics[scale=0.3]{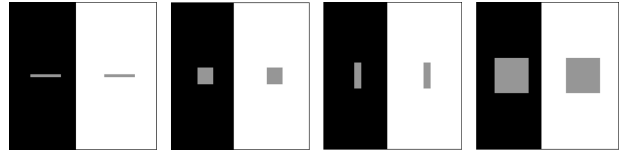}
%\vspace{-0.2cm}
\caption{Different types of SBC illusions selected as the comparator. The intensity level of the target is always 150. However the length and width of the target varies widely.}
%\vspace{-0.2cm}
\label{fig:exp_setup_supp}
\end{figure}
%%%%%%%%%%%%%%%%%%%%%%%%%%%%%%%%%%%%%%%%%%%%%%%%%%%%%%%%%%%%%%%%%%%%%%%%%%%%%%%%%%%%%%%%%%%%%%%%%%

%%%%%%%%%%%%%%%%%%%%%%%%%%%%%%%%%%%%%%%%%%%%%%%%%%%%%%%%%%%%%%%%%%%%%%%%%%%%%%%%%%%%%%%%%%%%%%%%%%
\begin{figure*}[!t]
\centering
\includegraphics[scale=0.35]{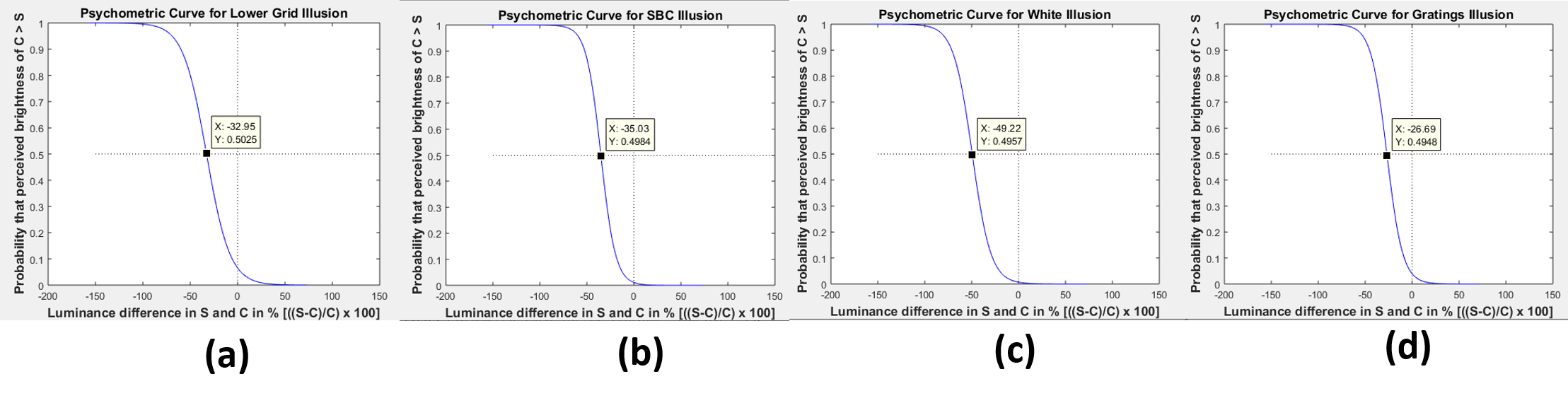}
%\vspace{-0.5cm}
\caption{Psychometric curve for (a)Lower grid, (b) SBC, (c) White illusion, (d) Grating illusion taken from the generated dataset.
This indicates the probability that the perception agrees with the reality as a function of the real difference in luminance between the standard (S) and the comparator (C).
The illusory reduction for Grating, White illusion, SBC, Lower grid are 26.69, 49.22, 35.03 and 32.95 respectively. }
%\vspace{-0.2cm}
\label{fig:psychometric_curve_supp}
\end{figure*}
%%%%%%%%%%%%%%%%%%%%%%%%%%%%%%%%%%%%%%%%%%%%%%%%%%%%%%%

\subsection{Materials and Methods}

The experimental arrangements are designed identical to that described in Mitra et al.~\cite{mitra2018adaptive} and Shi et al.~\cite{shi2013effect}. A chin rest is placed at a distance of d cm from a linearized video monitor (HP Compaq LE 2002X with resolution 1,024 X 1,024 pixels) and the subject’s head is stabilized on this chin rest. The value of d is chosen as 57 cm, because it may be shown from simple trigonometry that at such a distance an image of width (or length) of one cm. subtends a visual angle of approximately 1 degree. Subjects binocularly view the visual presentations, keeping their heads fixed on the chin rest.
In these experiments, the subjects visually compare the brightness or lightness of two targets, namely the standard and the comparator. We conduct a two-alternative forced-choice experiment for lightness discrimination between the comparator and the standard and measured the illusory enhancement or illusory reduction. The subjects are instructed to give his/her judgment by pressing the key marked ONE when the comparator appeared to be lighter than the standard, otherwise key marked as TWO was to be pressed.
The comparator is an intensity based illusion generated statistically and it is selected individually from four types of illusions namely Simultaneous Brightness Contrast (SBC), White Illusion (WI), Gratings and Lower Grid illusions. 
In one set of experiment, the same type of the illusion, say SBC illusion is selected as the comparator. One such set of SBC illusion is shown in Fig.~\ref{fig:exp_setup_supp}. The intensity level of the target remains always 150, however it appears either lighter or, darker depending on the background intensity. In all the trials, the darker target is selected as the comparator, whose intensity is compared with the standard.

In our experiments, the standard is a target, containing a number of segments. Each segment could be distinguished from the other by its intensity of gray value. For example, our striped standard was divided into 11  segments of varying intensity values values 13, 36, 59, 82, 105, 128, 150, 173, 196, 219 and 242. These values are kept fixed during the entire experiment, although the order of appearance of these 11 segments within the standard are scrambled pseudo-randomly. So the appearance of the standard varies from trial to trial.

The screen design for two-alternate forced-choice experiment is shown in Fig. 6 (in the paper) for White illusion and SBC respectively.
In each trial, the subject has been first instructed to fix attention on a central red cross (1 deg within a 3.5 deg fixation window). 
Within one second the comparator and standard appears on the screen simultaneously, one of them centered at 7 deg to the left of the red cross and the other centered at 7 deg to the right of the red cross. Two vertical red lines are introduced on the comparator and standard at the same height of the red cross, which denotes the specific area of interest of them to be compared. 
The comparator and standard exists on the computer screen for about 3 seconds before disappearing. Subjects need not had to wait to give their judgments till the stimuli disappeared from the display. Red lines indicate the region of interest in the comparator to be judged against the nearest segment of the standard. If the comparator appears to be lighter than the standard, subjects have to press key number ONE, otherwise they have to press key number TWO. During one set of experiment (say for Simultaneous Brightness Contrast illusion), both the comparator (varying its width and length) and standard (by scrambling 11 intensity levels) are changed from trial to trial.

The perceived difference of brightness between the comparator and standard depends on the actual difference of brightness and also on the psychophysical response of the subject. To keep the subjects unbiased and attentive, various parameters were changed randomly from trial to trial. As an example, we can state that the comparator can appear to the left or right of the red cross, the intensity of the standard appearing at the same height of the comparator can vary widely etc.
Perceived difference of brightness between the comparator and the standard depends on the actual difference of brightness between those and also the psychophysical effect on the subject. If the actual difference in the brightness of the co-occurring comparator and the standard is zero, the apparent perceived difference is then entirely due to the psychophysical effect.

\subsection{Experimental Results}

The purpose of psychophysical experiment is to measure \textit{Point of Subjective Equality} (PSE)~\cite{mitra2018adaptive}, which is a measure involving actual intensities of the comparator and standard, when these appear to be same to the subject. 
In our experiments, the illusory enhancement or illusory decrement are measured at the point of subjective equality from the different sets of psychometric curves drawn for four different illusions. 
As always we measure the illusory effect of the darker target, we can calculate the average illusory decrement for each type of illusion under consideration.

Two sets of psychometric curves are presented in the paper. They represent the psychometric curves for Simultaneous Brightness Contrast (SBC) illusion and White illusion. 
We have performed similar experiments for other variants of illusions present in the dataset.
The psychometric curves for SBC, White illusion, gratings illusion and lower grid illusion are provided in Fig.~\ref{fig:psychometric_curve_supp}. Illusory decrement indicates the quantitative measure of perceived darkness. For example, given the standard grayscale value to be 150, illusory reduction of 32.95 indicates that the target is perceived to be grayscale value of 117.05 and looks darker. The illusory reduction of the same intensity comparator of grayscale value 150 for White illusion is maximum (49.22), for gratings illusion is minimum (27.11), and it lies within this maximum and minimum limit for SBC (35.03) and Lower-grid illusion (32.95) as shown in Fig.~\ref{fig:psychometric_curve_supp}. These psycophysical experiments statistically validate the dataset. 
%%%%%%%%%%%%%%%%%%%%%%%%%%%%%%%%%%%%%%%%%%%%%%%%%%%%%%%%%%%%%%%%%%%%%%%%%%%%%%%%%%%%%%%%%%%%%%%%%%
\begin{figure}[!t]
\centering
\includegraphics[scale=0.3]{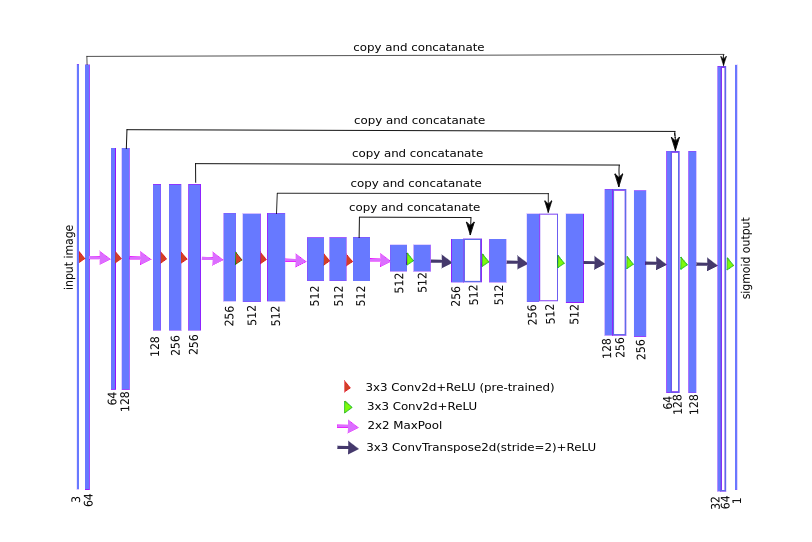}
%\vspace{-0.2cm}
\caption{UNet architecture~\cite{ronneberger2015u}}
%\vspace{-0.2cm}
\label{fig:unet}
\end{figure}
%%%%%%%%%%%%%%%%%%%%%%%%%%%%%%%%%%%%%%%%%%%%%%%%%%%%%%

%%%%%%%%%%%%%%%%%%%%%%%%%%%%%%%%%%%%%%%%%%%%%%%%
\begin{figure}[!t]
\centering
\includegraphics[scale=0.35]{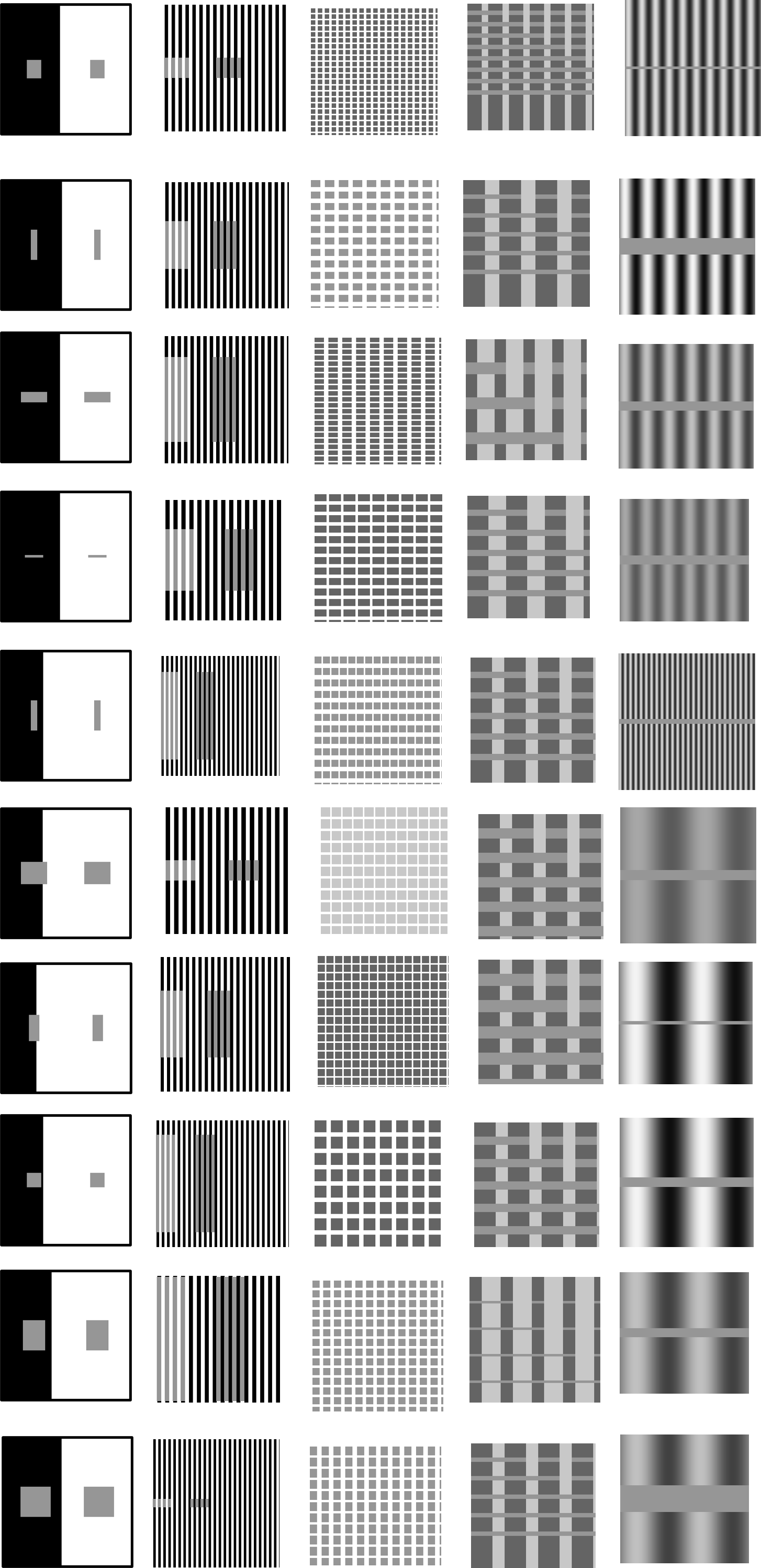}
%\includegraphics[scale=0.35]{figures/ill_ex_sup.png}
%\vspace{-0.2cm}
\caption{Illusion examples: SBC (first column), White illusion (second column), Hermann grid (third column), grid illusion (fourth column) and grating illusion (fifth column) with varying grayscale values, scales and frequencies. These images are taken from our generated dataset to show the variety of scales, contrast and orientation.}
%\vspace{-0.2cm}
\label{fig:ill_example}
\end{figure}
%%%%%%%%%%%%%%%%%%%%%%%%%%%%%%%%%%%%%%%%%%%%%%%%%%%%%%%%%%%%%%%%%%%%%%%%%%%%%%%%%%%%%%%%%%%%%%%%%%

%%%%%%%%%%%%%%%%%%%%%%%%%%%%%%%%%%%%%%%%%%%%%%%%%%%%%%%%%%%%%%%%%%%%%%%%%%%%%%%%%%%%%%%%%%%%%%%%%%
\begin{figure}[!t]
\centering
\includegraphics[scale=0.3]{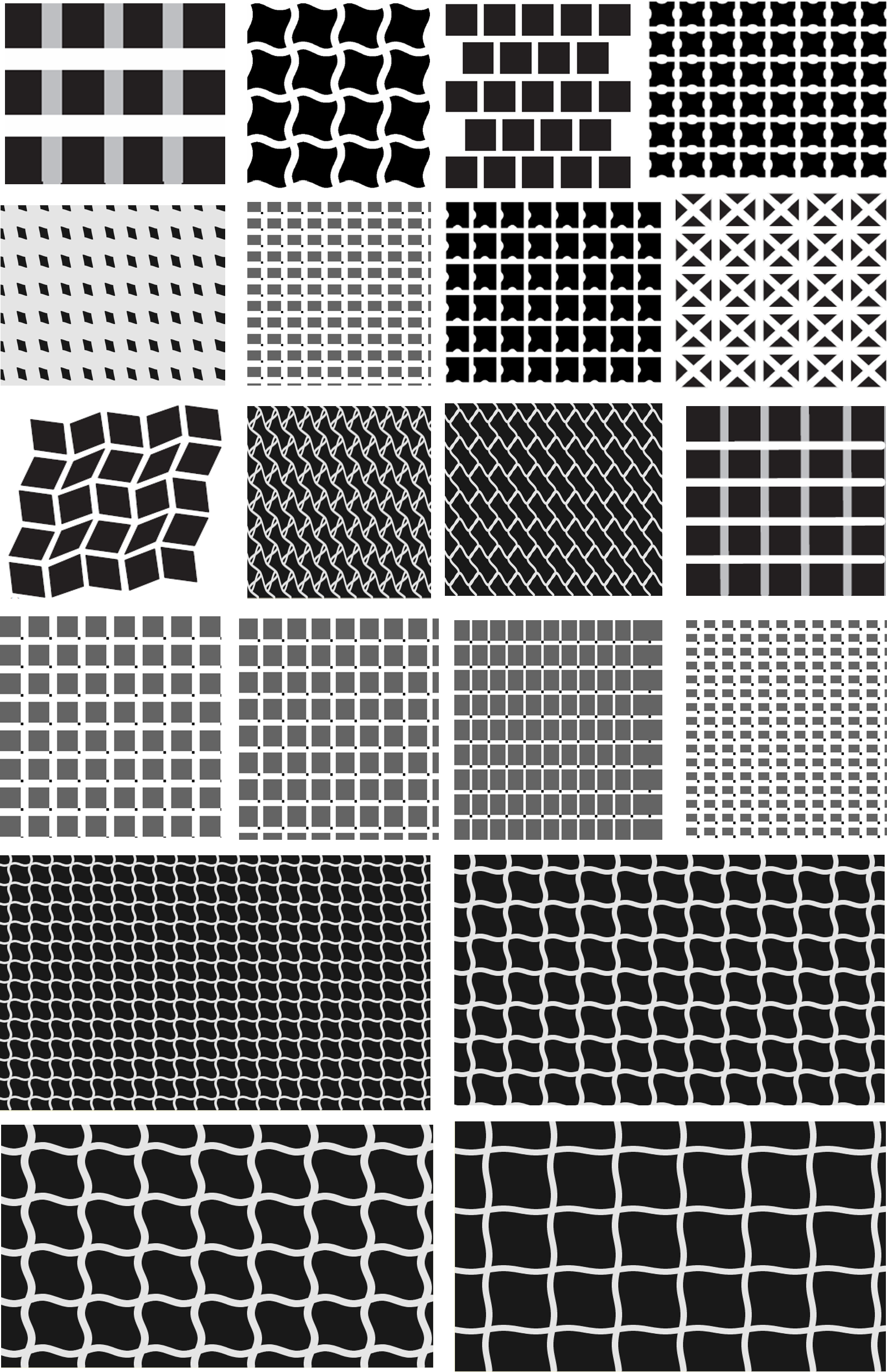}
%\vspace{-0.2cm}
\caption{Non-illusion examples generated by us in the dataset by introducing non-linearity, dot-insertion~\cite{bakshi2020tiny,geier2008straightness} etc.}
%\vspace{-0.2cm}
\label{fig:non_ill_example}
\end{figure}
%%%%%%%%%%%%%%%%%%%%%%%%%%%%%%%%%%%%%%%%%%%%%%%%%%%%%%%%%%%%%%%%%%%%%%%%%%%%%%%%%%%%%%%%%%%%%%%%%%

\section{Dataset}

\subsection{Dataset generation}
The dataset is generated by creating standard brightness illusions (i.e., SBC, White illusion, Hermann grid, upper and lower grid, grating illusion) with varying patch size, frequency, grayscale values as shown in Fig.~\ref{fig:ill_example}.
1149 non-illusion variants are also generated by utilizing several well-known results~\cite{bakshi2020tiny,geier2008straightness} as shown in Fig.~\ref{fig:non_ill_example}.
The dataset will be made public.

\section{Network details}
\subsection{Illuison identificaion}

For illusion identification, we finetune a ResNet-18 model. ResNet-18 is a convolutional neural network that is 18-layer deep~\cite{he2016deep}. The training has been done with stochastic gradient descent with learning rate of 0.01 and momentum of 0.9. 
We have also tried other variants like AlexNet~\cite{krizhevsky2012imagenet}(convolutional neural network containing 8 layers) and SqueezeNet~\cite{iandola2016squeezenet}(10 layer convolutional neural network with fewer parameters) to perform illusion identification and classification. The result is shown in Table 1 (in the paper).

\subsection{Illusion localization}

\begin{figure*}[!t]
\centering
\includegraphics[scale=0.5]{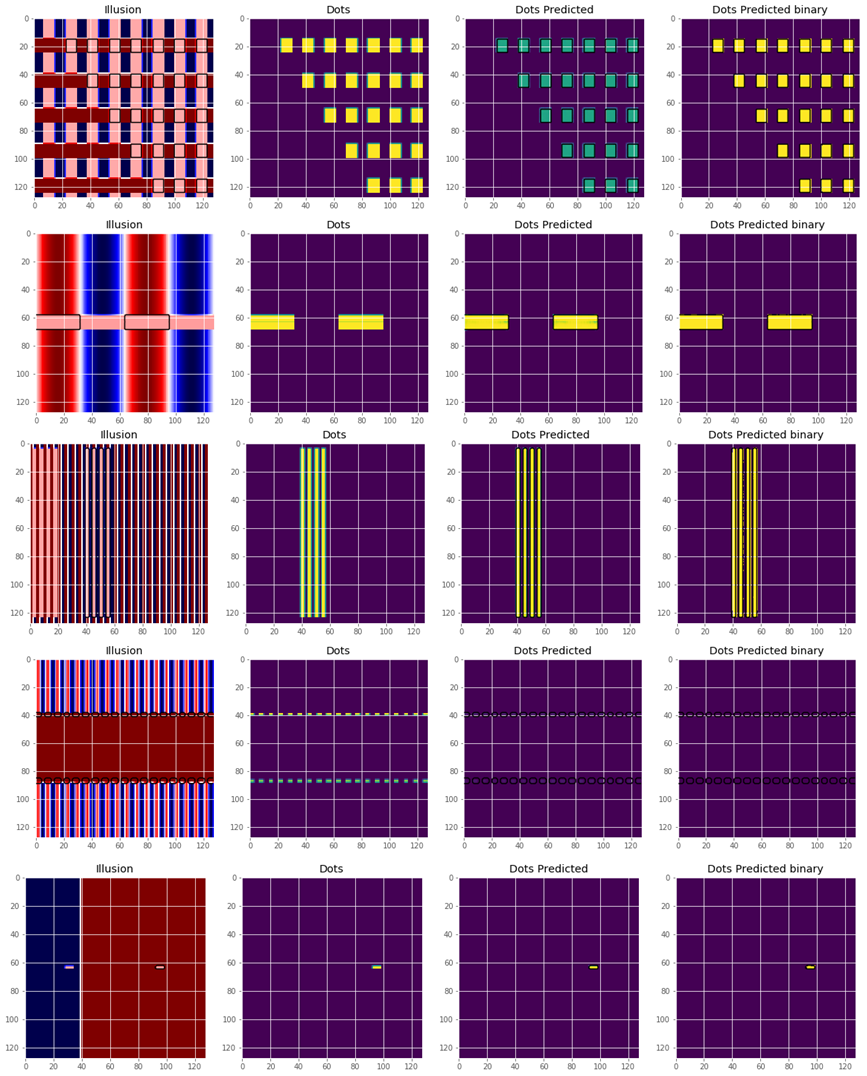}
%\vspace{-0.2cm}
\caption{Network response for illusion localization on test set: (a) test images, (b) corresponding groundtruth, (c) network response, (d) Binary response of the network by thresholding at 0.22. The trained network can localize perceived darkness quite efficiently.}
%\vspace{-0.2cm}
\label{fig:illustration}
\end{figure*}

For illusion localization, we finetune a UNet architecture to segment out the perceived darkness. 
The UNet was designed by Ronneberger et al.~\cite{ronneberger2015u} for bio-medical image segmentation. The architecture contains two paths. The first path is the contraction path (also called as the encoder, refer to Fig.~\ref{fig:unet}) which is used to capture the context in the image.
The second path is the symmetric expanding path (also called as the decoder, refer to Fig.~\ref{fig:unet}) which is used to enable precise localization using transposed convolutions.
The exact architecture is shown in Fig.~\ref{fig:unet}.
The experimental details are given in Table 3 (in the main paper).
We consider a weighted combination of MSE loss ($L_{MSE}$ and SSIM loss ($L_{SSIM}$) as the loss function, $L = \alpha L_{MSE} + \beta L_{SSIM}$, where $L_{MSE}$ is mean squared error loss and $L_{SSIM}$~\cite{zhao2016loss} is the SSIM loss. We experimentally set the parameters, $\alpha = 0.4$, $\beta = 0.6$, which produced the best performance as shown in Tab.3 (in the main paper).
Adam optimizer is used for training with batch size of 32 and a learning rate of 0.001.
Some of the qualitative results on test images are shown in Fig.~\ref{fig:illustration}.

%%%%%%%%%%%%%%%%%%%%%%%%%%%%%%%%%%%%%%%%%%%%%%%%%%%%%%%%%%%%%%%%%%%%%%%%%%%%%%%%%%%%%%%%%%%%%%

%%%%%%%%%%%%%%%%%%%%%%%%%%%%%%%%%%%%%%%%%%%%%%%%%%%%%%%%%%%%%%%%%%%%%%%%%%%%%%%%%%%%%%%%%%%%%%%%%%
%%%%%%%%%%%%%%%%%%%%%%%%%%%%%%%%%%%%%%%%%%%%%%%%%%%%%%%%%%%%%%%%%%%%%%%%%%%%%%%%%%%%%%%%%%%%%%%%%%

%%%%%%%%%%%%%%%%%%%%%%%%%%%%%%%%%%%%%%%%%%%%%%%%%%%%%%%%%%%%%%%%%%%%%%%%%%%%%%%%%%%%%%%%%%%%%%%%%%
\begin{figure}[!t]
\centering
\includegraphics[scale=0.3]{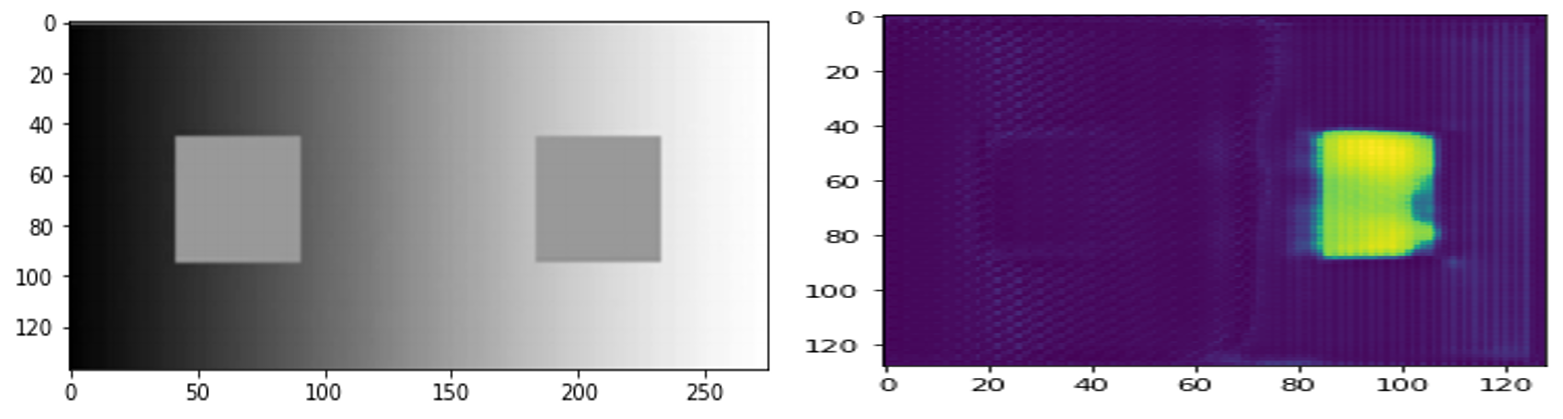}
%\vspace{-0.2cm}
\caption{Network response for SBC with luminance gradient: illusion image (left), network response (right). Instead of binary background in SBC, the background has luminance gradient. The model still identifies the darker patch.}
%\vspace{-0.2cm}
\label{fig:step_SBC}
\end{figure}
%%%%%%%%%%%%%%%%%%%%%%%%%%%%%%%%%%%%%%%%%%%%%%%%%%%%%%%%%

%%%%%%%%%%%%%%%%%%%%%%%%%%%%%%%%%%%%%%%%%%%%%%%%%%%%%%%%%%%%%%%%%%%%%%%%%%%%%%%%%%%%%%%%%%%%%%%%%%
\begin{figure}[!t]
\centering
\includegraphics[scale=0.38]{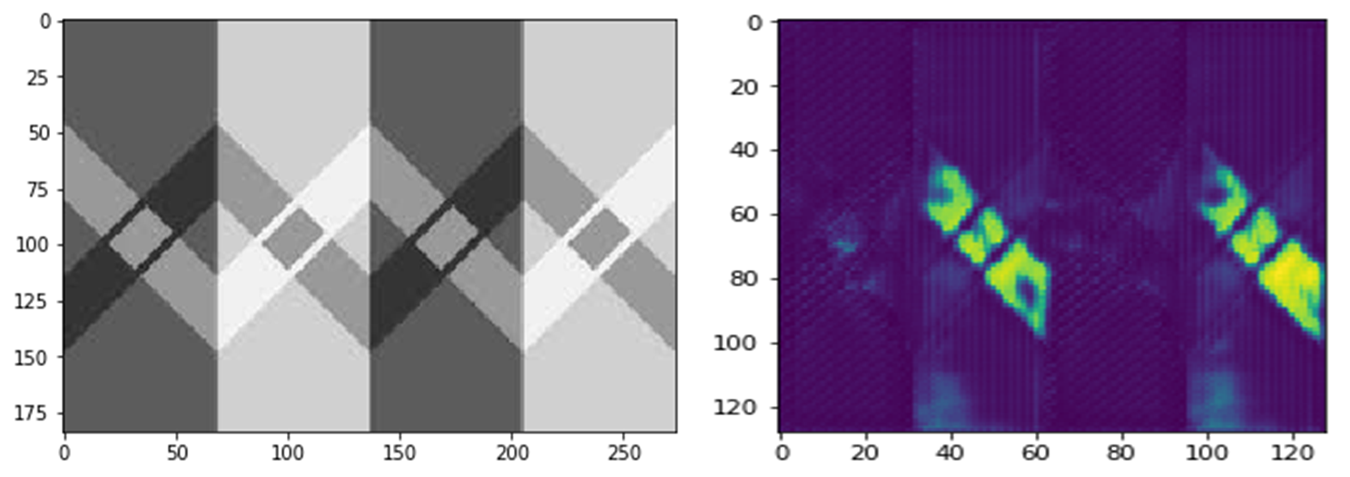}
%\vspace{-0.2cm}
\caption{Network response for criss-cross brightness illusion: illusion image (left), network response (right). The darker patches are identified by the model.}
%\vspace{-0.2cm}
\label{fig:criss_cross}
\end{figure}
%%%%%%%%%%%%%%%%%%%%%%%%%%%%%%%%%%%%%%%%%%%%%%%%%%%%%%%%%

%%%%%%%%%%%%%%%%%%%%%%%%%%%%%%%%%%%%%%%%%%%%%%%%%%%%%%%%%%%%%%%%%%%%%%%%%%%%%%%%%%%%%%%%%%%%%%%%%%
\begin{figure}[!t]
\centering
\includegraphics[scale=0.66]{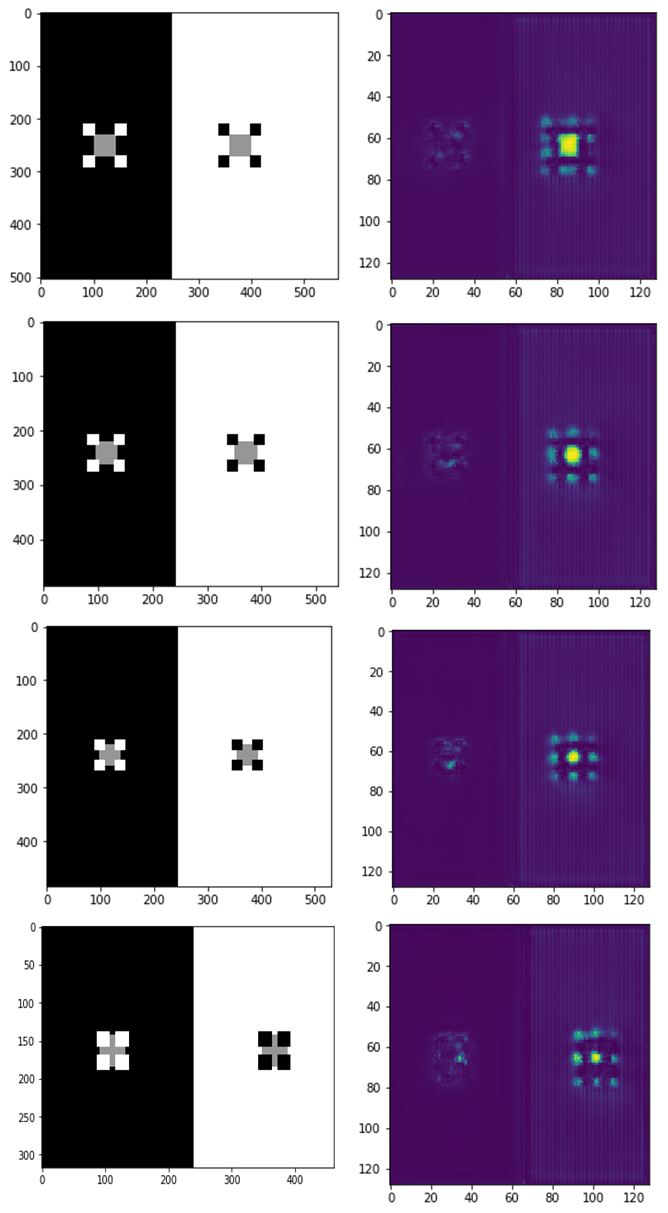}
%\vspace{-0.2cm}
\caption{Network response for todorovic illusion: todorovic illusion (left), network response (right). As the gray patch occludes by surrounding, the illusory effect is more focused in the center.}
%\vspace{-0.2cm}
\label{fig:todorovic}
\end{figure}
%%%%%%%%%%%%%%%%%%%%%%%%%%%%%%%%%%%%%%%%%%%%%%%%%%%%%%%%%

%%%%%%%%%%%%%%%%%%%%%%%%%%%%%%%%%%%%%%%%%%%%%%%%%%%%%%%%%%%%%%%%%%%%%%%%%%%%%%%%%%%%%%%%%%%%%%%%%%
\begin{figure}[!t]
\centering
\includegraphics[scale=0.33]{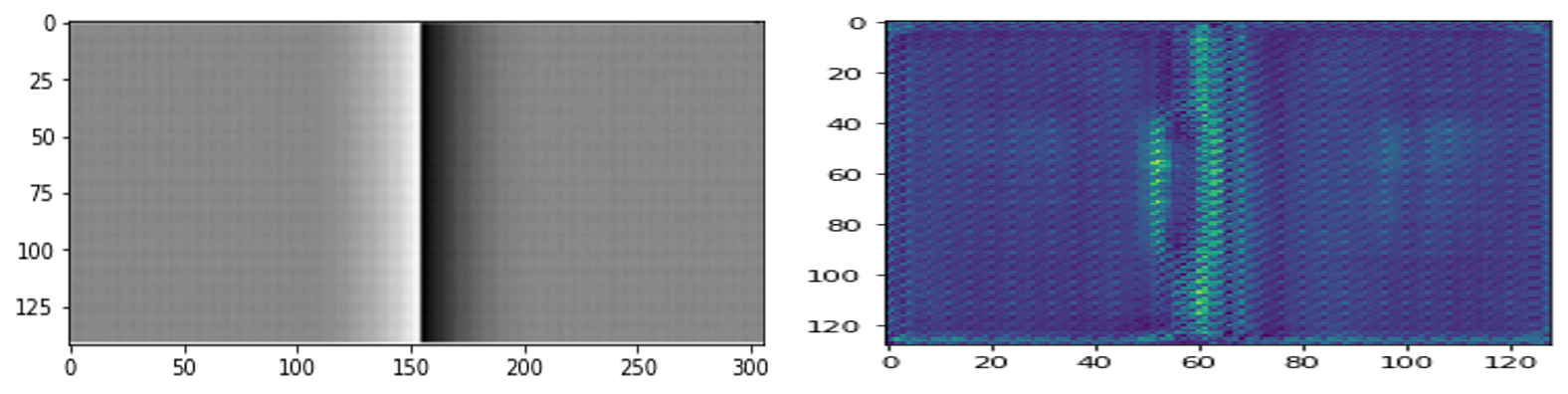}
%\vspace{-0.2cm}
\caption{Network response for cornsweet illusion: cornsweet illusion (left), network response (right). The model correctly identifies the darker patch at the right.}
%\vspace{-0.2cm}
\label{fig:cornsweet}
\end{figure}
%%%%%%%%%%%%%%%%%%%%%%%%%%%%%%%%%%%%%%%%%%%%%%%%%%%%%%%%%

%%%%%%%%%%%%%%%%%%%%%%%%%%%%%%%%%%%%%%%%%%%%%%%%%%%%%%%%%%%%%%%%%%%%%%%%%%%%%%%%%%%%%%%%%%%%%%%%%%
\begin{figure}[!t]
\centering
\includegraphics[scale=0.3]{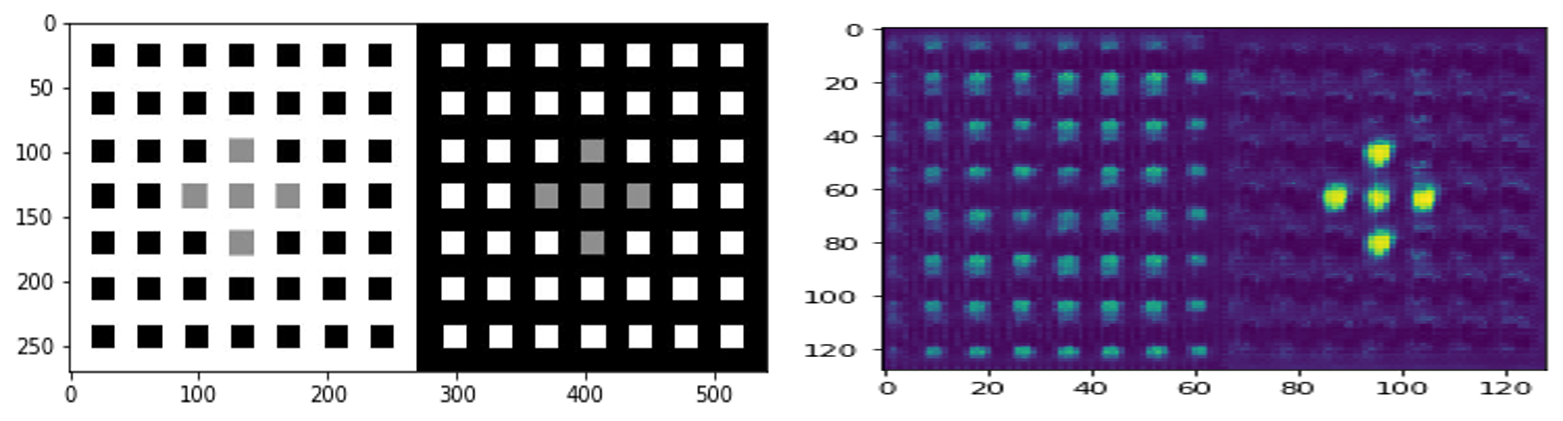}
%\vspace{-0.2cm}
\caption{Network response for dungeon illusion: dungeon illusion (left), network response (right). The gray patches in the right looks darker, which the model correctly predicts. It also predicts Hermann grid blobs in the left side.}
%\vspace{-0.2cm}
\label{fig:bressan_1}
\end{figure}
%%%%%%%%%%%%%%%%%%%%%%%%%%%%%%%%%%%%%%%%%%%%%%%%%%%%%%%%%

% %%%%%%%%%%%%%%%%%%%%%%%%%%%%%%%%%%%%%%%%%%%%%%%%%%%%%%%%%%%%%%%%%%%%%%%%%%%%%%%%%%%%%%%%%%%%%%%%%%
% \begin{figure}[!t]
% \centering
% \includegraphics[scale=0.3]{figures/bressan_2.png}
% %\vspace{-0.2cm}
% \caption{Network response for dungeon illusion variant 2: dungeon illusion variant 2 (left), network response (right).}
% %\vspace{-0.2cm}
% \label{fig:bressen_2}
% \end{figure}
% %%%%%%%%%%%%%%%%%%%%%%%%%%%%%%%%%%%%%%%%%%%%%%%%%%%%%%%%%

%%%%%%%%%%%%%%%%%%%%%%%%%%%%%%%%%%%%%%%%%%%%%%%%%%%%%%%%%%%%%%%%%%%%%%%%%%%%%%%%%%%%%%%%%%%%%%%%%%
\begin{figure}[!t]
\centering
\includegraphics[scale=0.45]{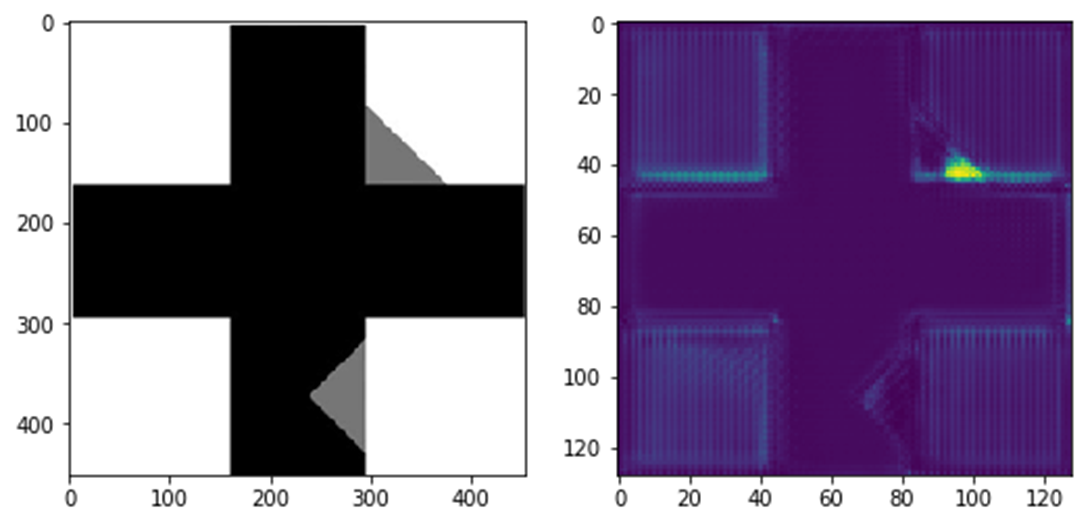}
%\vspace{-0.2cm}
\caption{Network response for Benary's cross: Benary's cross (left), network response (right). The upper right gray patch seems to look darker, which the model correctly predicts.}
%\vspace{-0.2cm}
\label{fig:benary_cross}
\end{figure}
%%%%%%%%%%%%%%%%%%%%%%%%%%%%%%%%%%%%%%%%%%%%%%%%%%%%%%%%%

\section{Qualitative results on unseen illusions}

In this section, we verify the efficacy of our proposed data-driven approach on some more well-known illusions, which have not been used during training.
Qualitative network response for variants of dungeon illusion~\cite{bressan2001explaining} (Fig.~\ref{fig:bressan_1}), SBC with luminance gradient~\cite{poom2020influences} (Fig.~\ref{fig:step_SBC}), criss-cross brightness illusion~\cite{bressan2001explaining}(Fig.~\ref{fig:criss_cross}), Todorovic illusion~\cite{todorovic1997lightness}(Fig.~\ref{fig:todorovic}), benary's cross~\cite{gilchrist2014gestalt} (Fig.~\ref{fig:benary_cross}) and cornsweet illusion~\cite{poom2020influences} (Fig.~\ref{fig:cornsweet}) are shown here. These results suggest that the trained model can also correctly predict the perceived darkness on these type of unseen illusions.

\begin{table}[!t]
\centering
\scalebox{0.7}{
\begin{tabular}{|c|c|c|}
\hline
\textbf{Illusion} & \textbf{Accuracy} & \textbf{mIoU} \\ \hline
Hermann grid      & 0.6905            & 0.43          \\ \hline
Induced grating   & 0.9206            & 0.49          \\ \hline
Lower grid        & 0.7991            & 0.43          \\ \hline
Upper grid        & 0.8546            & 0.44          \\ \hline
SBC               & 0.8805            & 0.48          \\ \hline
White             & 0.9745            & 0.5           \\ \hline
\end{tabular}}
\caption{ \small Transferring illusions across other brightness illusions: testing on particular illusion as mentioned in the rows (from our dataset) while training the segmentation model for dark region localization on all other types except that particular illusion.}
\label{tab:ill_transfer}
\end{table}

\section{ Generalization to unseen illusions.}

To check whether the learned model generalizes among brightness illusions, we perform the following experiment. Among the five types of illusions, we train a segmentation network on four types of illusions and test on the unseen one. We show the results in Table.~\ref{tab:ill_transfer}. For example, we train on all the illusion images except the Hermann grid and test its generalization performance on the Hermann grid and observed illusion localization accuracy of 69.05\% with mIoU of 0.43 as shown in Table.~\ref{tab:ill_transfer}. We find that the learned segmentation model generalizes poorly when testing on the Hermann grid. However, for all other cases, the trained model transfers quite effectively for the other four types of illusions as shown in Table.~\ref{tab:ill_transfer}. We believe, as in the Herman grid, the illusory regions are not physically present and thus make the localization challenging.

Beyond these five types of illusion, we also evaluate of other popular unseen illusions such as Mach band illusion, non-linear Hermann grid, and Dungeon illusion. 

\begin{figure}[!t]
   \begin{minipage}{0.48\textwidth}
    \centering
    \includegraphics[scale=0.25]{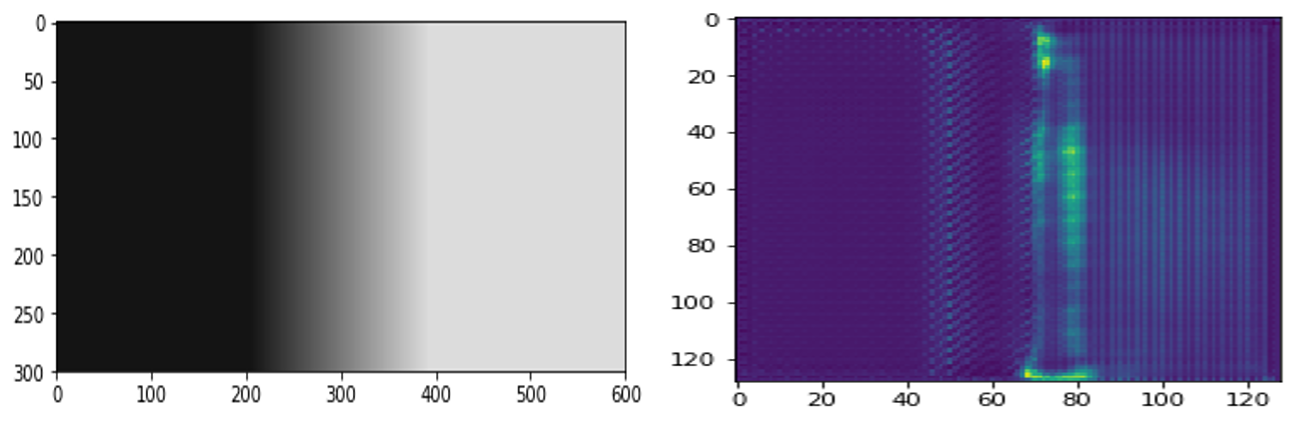}
    \vspace{-0.1cm}
    \caption{Mach band illusion (left) and corresponding learned model output (right). Two dark parallel dark patch indicates Mach band.}
    \vspace{-0.2cm}
    \label{fig:mach_band}
   \end{minipage}\hfill
   \begin{minipage}{0.48\textwidth}
    \centering
    \includegraphics[scale=0.23]{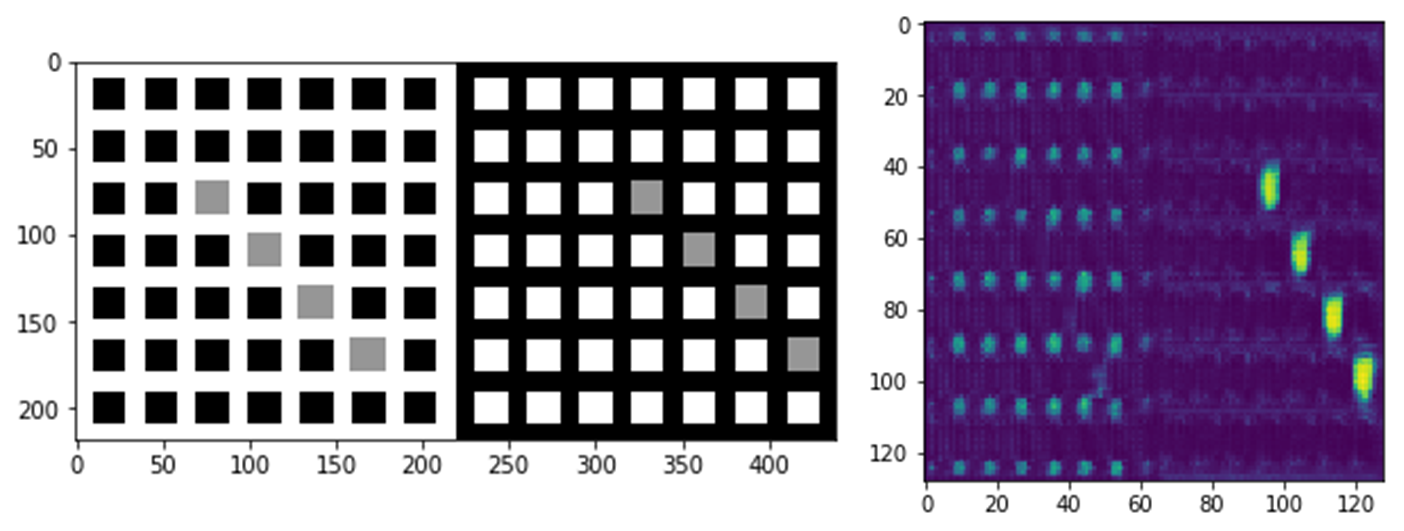}
    \vspace{-0.3cm}
    \caption{Dungeon illusion and corresponding response form the network: Right region seems to be darker than left ones, however they are exactly the same grayscale level.}
    \vspace{-0.2cm}
    \label{fig:dungeon}
   \end{minipage}
\end{figure}

\textbf{Mach band illusion.}
Mach bands are apparent bright and dark lines occurring at the border between objects with different optical densities, contrast levels, or luminances~\cite{buckle2013now}.
Fig.~\ref{fig:mach_band} indicates that our learned model can detect Mach bands. Note, since we have trained to identify the darker patch, the learned model can identify the bright Mach band by displaying two surrounding parallel darker regions as shown in Fig.~\ref{fig:mach_band}.
% Mach band has several applications in medical imaging to identify lesions, fracture in radiology images as we will be discussing in Sec.~\ref{sec:application}.

%%%%%%%%%%%%%%%%%%%%%%%%%%%%%%%%%%%%%%%%%%%%%%%%%%%%%%%%%%%%%%%%%%%%%%%%%%%%%%%%%%%%%%%%%%%%%%%%%

\section{Illusion vs natural image.}

We test illusion vs natural image (from the Caltech101 dataset~\cite{fei2004learning}) classification and achieve a high test accuracy of 99.98 \% in only 5 epochs.
% Thus, to make the problem non-trivial, we generate the non-illusion images by perturbing the illusions by following the common psychophysics techniques~\cite{bakshi2020tiny, geier2008straightness}. Note that the resulting non-illusion images are more visually similar to the illusion images than natural images.

\section{Test on illusions in natural images.} 

\begin{figure}[!t]
\centering
\vspace{-0.2cm}
\includegraphics[scale=0.22]{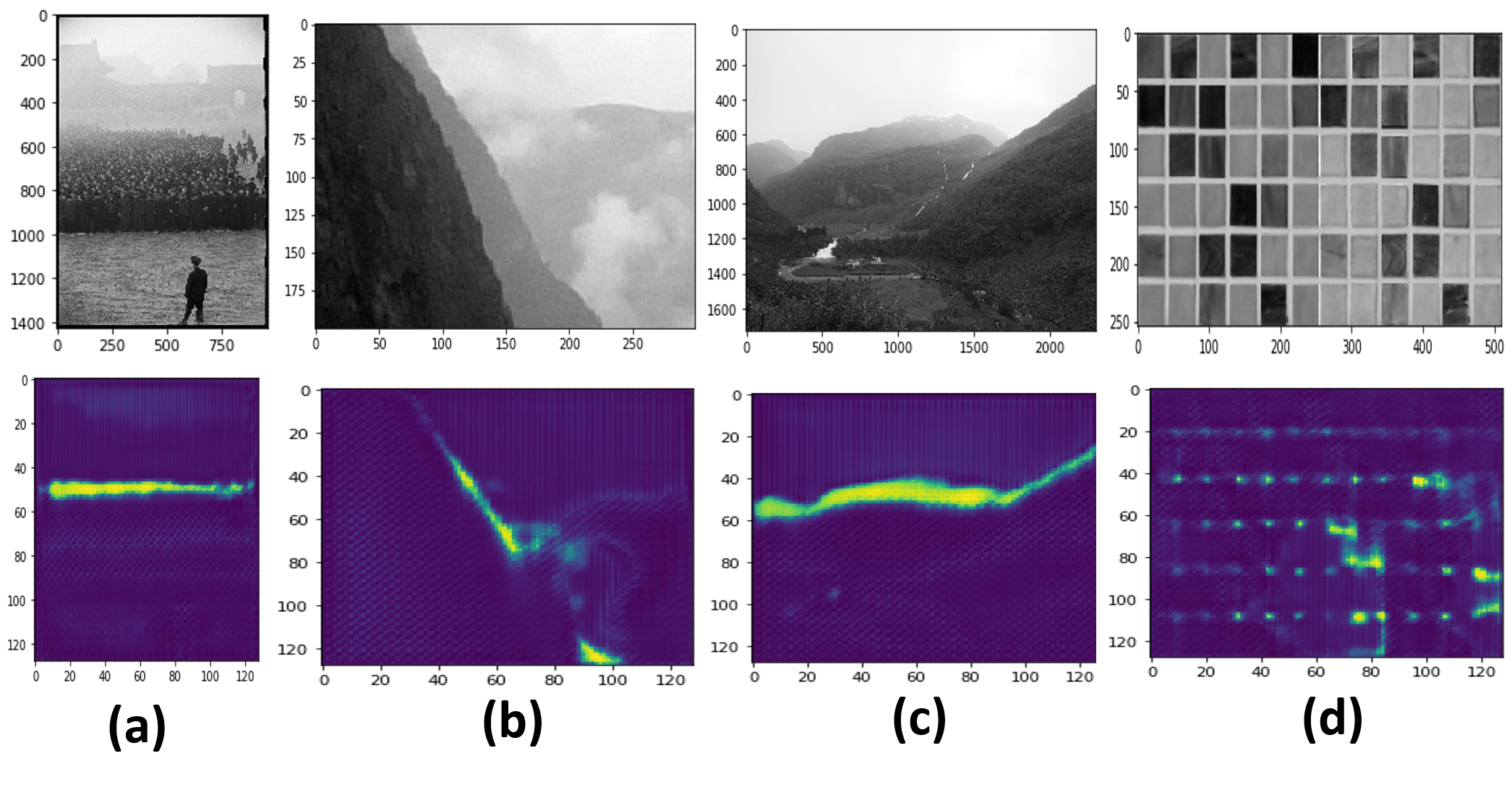}
\vspace{-0.2cm}
\caption{Model response (lower row) on natural images (upper row): (a), (b), (c) natural images showing Mach band, (d) Illusory regions identified in illusions generated from natural image by ~\cite{hirsch2020color}.}
\vspace{-0.2cm}
\label{fig:nat_img_ill}
\end{figure}

We test on natural images with diverse lighting and shading information which are used in illusion studies, e.g., Cartier-Bresson stimulus, as shown in Fig.~\ref{fig:nat_img_ill}~\cite{blakeslee2012spatial}. Our model can identify the darker regions and Mach bands in natural images as shown in Fig.~\ref{fig:nat_img_ill}.
We have also tested our framework with some of the recent approach of generating illusions from natural images~\cite{hirsch2020color}. Fig.~\ref{fig:nat_img_ill}(d) shows that our approach can identify illusory regions in those generated illusions too.

%%%%%%%%%%%%%%%%%%%%%%%%%%%%%%%%%%%%%%%%%%%%%%%%%%%%%%%%%%%%%%%%%%%%%%%%%%%%%%%%%%%%%%%%%%%%%%%%%%
%%%%%%%%%%%%%%%%%%%%%%%%%%%%%%%%%%%%%%%%%%%%%%%%%%%%%%%%%%%%%%%%%%%%%%%%%%%%%%%%%%%%%%%%%%%%%%%%%%

\section{Network Analysis}

We investigate the feature maps learned by CNN models while identifying the illusions. We compute layer-wise attribution using GradCam~\cite{selvaraju2017grad} to analyze what features are encoded across the CNN layers. This approach, for a given target output, computes the contribution of the image regions for the final prediction across the layers. We use the Captum library~\cite{kokhlikyan2020captum} for computing the layer-wise attributions. We compare the layer-wise attributions of the ResNet model trained on natural images and trained on illusion images. 

\begin{figure}[!t]
\centering
%\vspace{-0.2cm}
\includegraphics[scale=0.265]{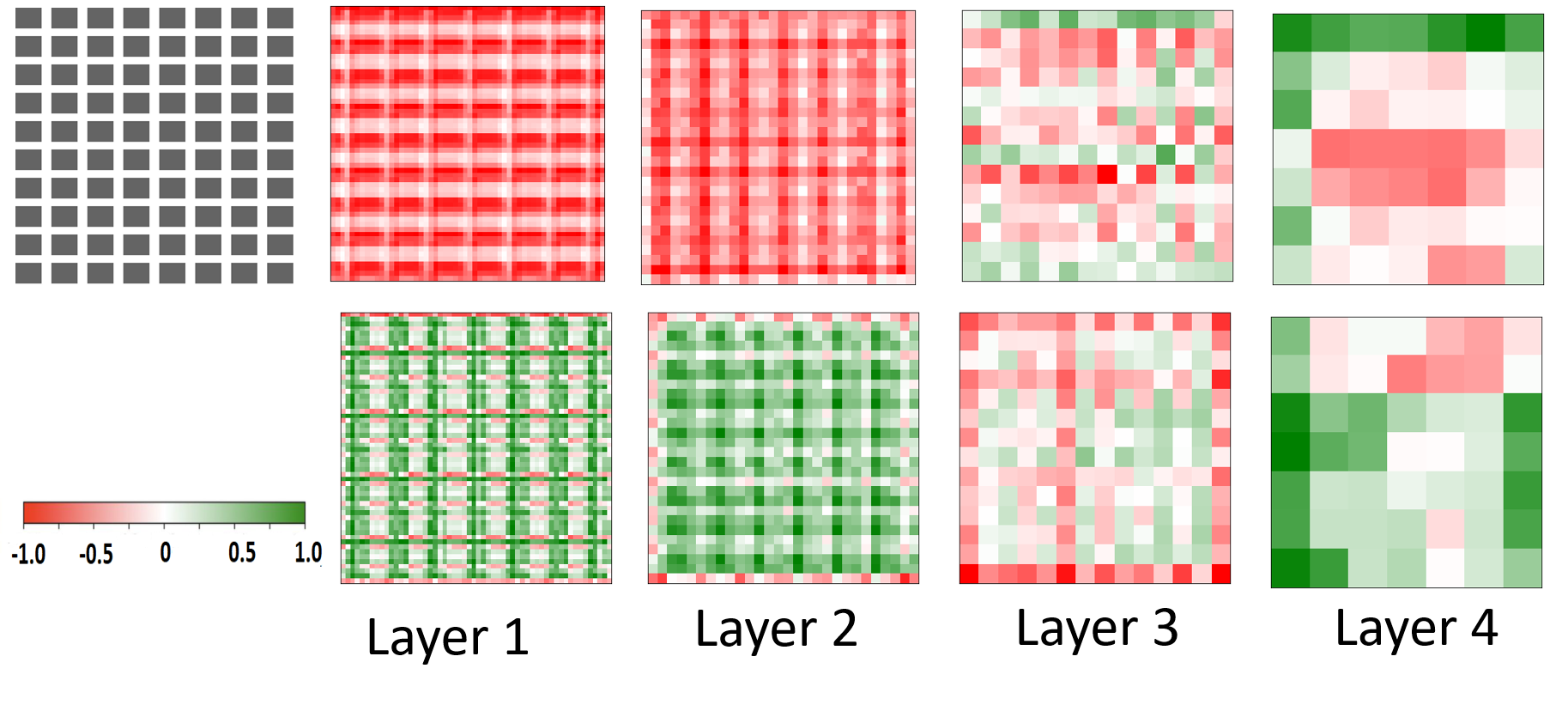}
%\vspace{-0.2cm}
\caption[\small]{Gradcam attributes of the ResNet18 layers for the Hermann grid illusion. Upper row: model trained with natural images using the ImageNet dataset. Lower row: model trained with illusions. The layer attributes are clearly different for both these cases. Also, illusion finetuned model seems to capture the darker regions or patches in the lower layer response.}
%\vspace{-0.2cm}
\label{fig:herman_grid_cam}
\end{figure}

\begin{figure}[!t]
\centering
%\vspace{-0.2cm}
\includegraphics[scale=0.25]{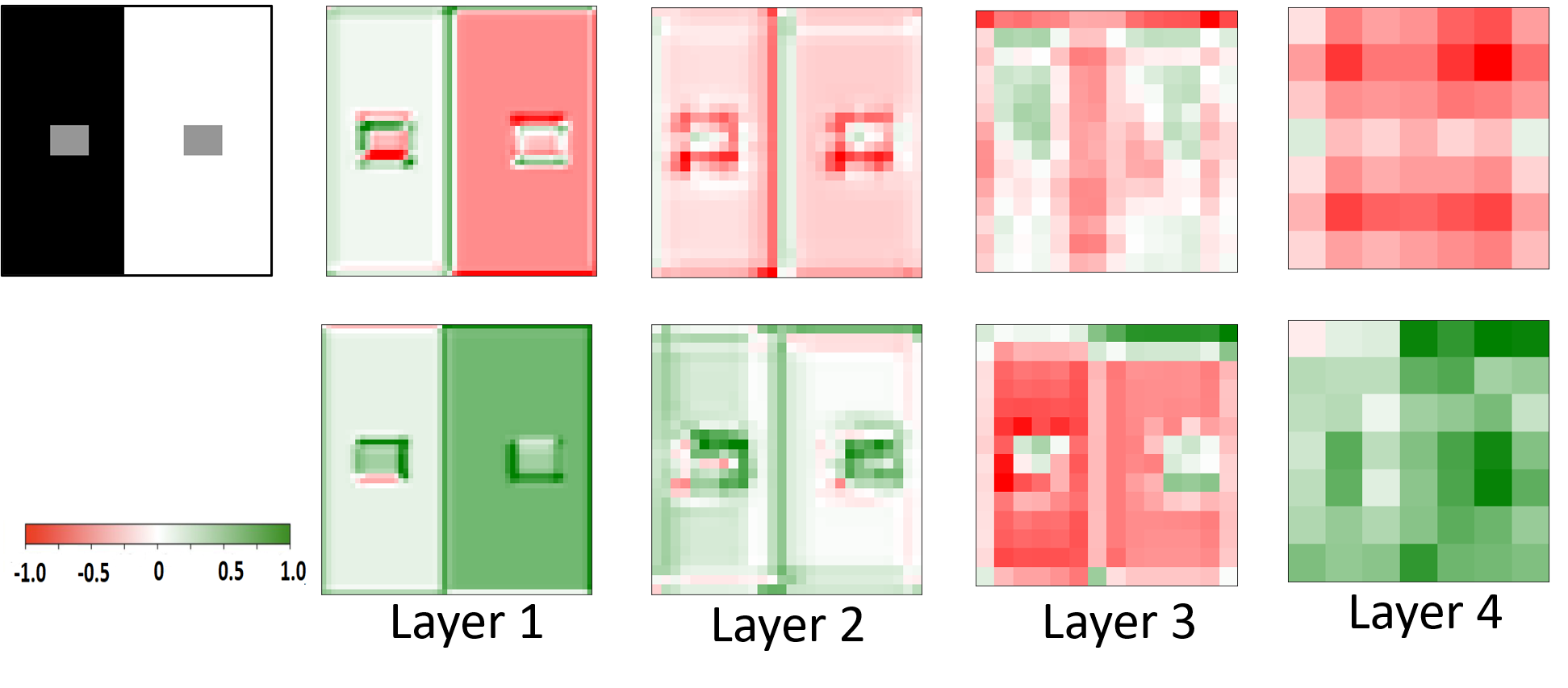}
%\vspace{-0.2cm}
\caption[\small]{Gradcam attributes of the ResNet18 layers for SBC. Upper row: model trained with natural images using the ImageNet dataset. Lower row: model trained with illusions. The layer attributes are clearly different for both these cases. Also, illusion finetuned model seems to capture the darker regions or patches in the lower layer response.}
%\vspace{-0.2cm}
\label{fig:sbc_gradcam}
\end{figure}

\begin{figure}[!t]
\centering
%\vspace{-0.2cm}
\includegraphics[scale=0.25]{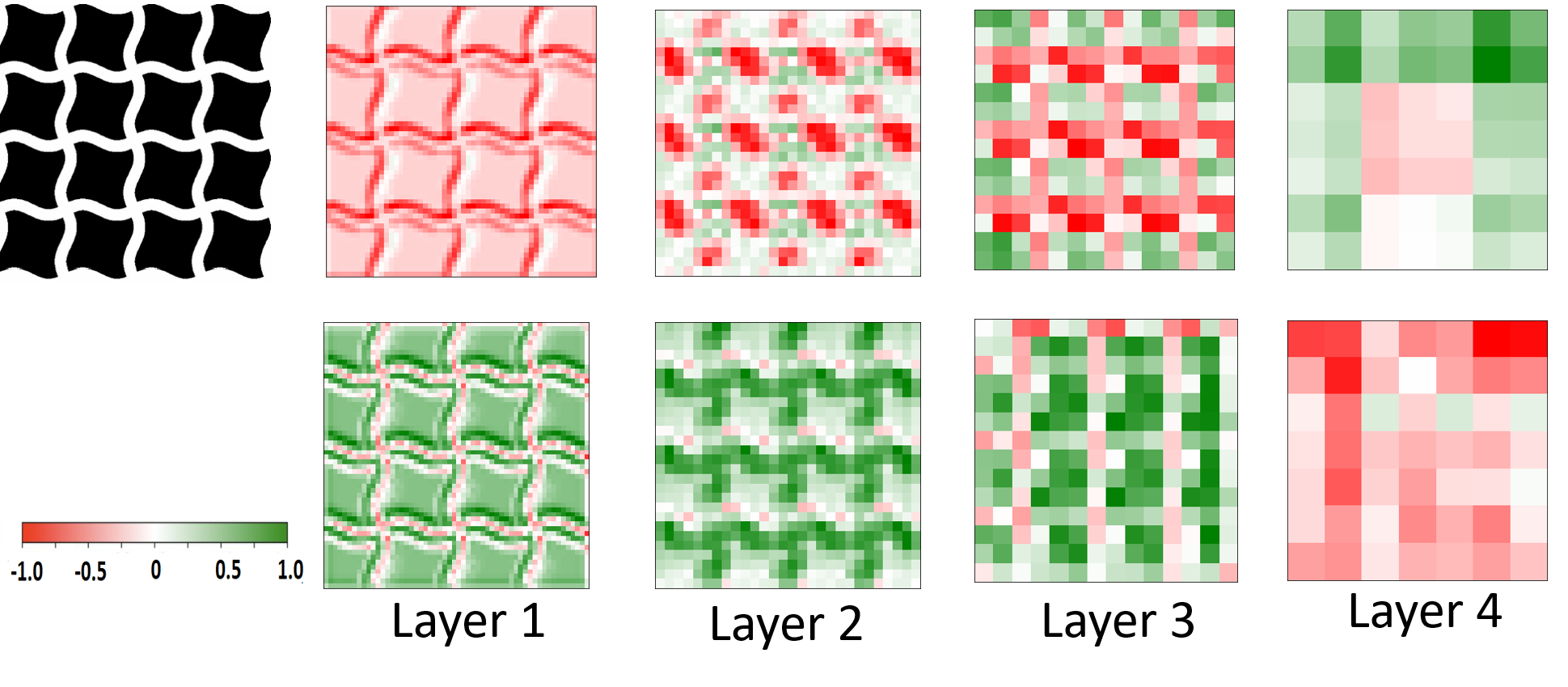}
%\vspace{-0.2cm}
\caption[\small]{Gradcam attributes of the ResNet18 layers for the non-illusion variant of Hermann grid. Upper row: model trained with natural images using the ImageNet dataset. Lower row: model trained with illusions.}
%\vspace{-0.2cm}
\label{fig:non_ill_gradcam}
\end{figure}

\begin{figure}[!t]
\centering
%\vspace{-0.2cm}
\includegraphics[scale=0.25]{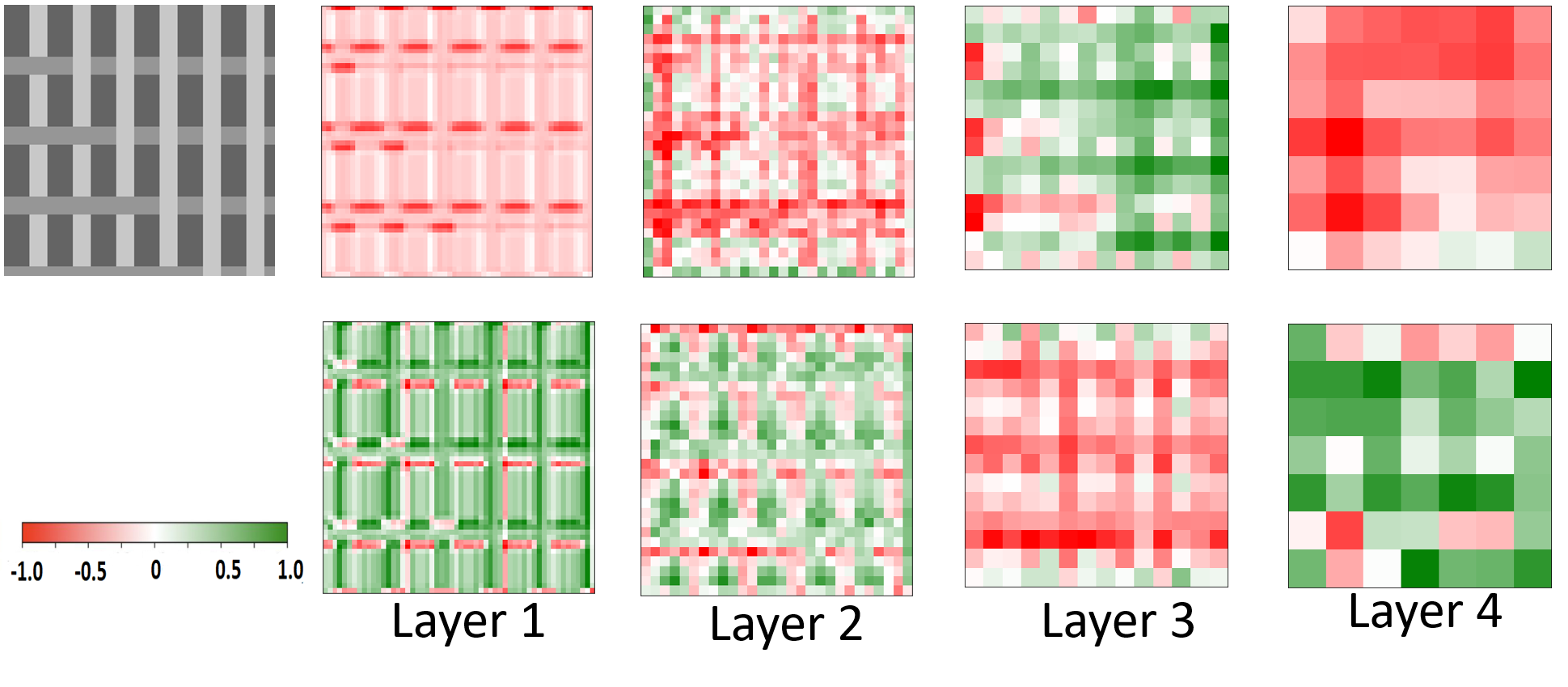}
%\vspace{-0.2cm}
\caption[\small]{Gradcam attributes of the ResNet18 layers for lower grid illusion. Upper row: model trained with natural images using the ImageNet dataset. Lower row: model trained with illusions.}
%\vspace{-0.2cm}
\label{fig:lower_grid_gradcam}
\end{figure}

In addition to the Gradcam attributes provided in the main paper, we are also providing some more examples of layer wise gradcam attributes for Hermann grid, SBC, non-illusion variant of Hermann grid and lower grid illusion in Fig.~\ref{fig:herman_grid_cam}, Fig.~\ref{fig:sbc_gradcam}, Fig.~\ref{fig:non_ill_gradcam} and Fig.~\ref{fig:lower_grid_gradcam} respectively.
As indicated in the main paper, there is a significant difference between the contribution of the image regions across the layers between the network trained with natural images (top row) and illusions (bottom row). 

In these examples also the lower layers tend to focus on low-level illusory features such as boundary between two regions with various intensities and the top layers learn more abstract features. The difference between the models trained with natural images and illusions are more prominent in the lower layers.

%%%%%%%%%%%%%%%%%%%%%%%%%%%%%%%%%%%%%%%%%%%%%%%%%%%%%%%%%%%%%%%%%%%%%%%%%%%%%%%%%%%%%%%%%%%%%%%%%%
%%%%%%%%%%%%%%%%%%%%%%%%%%%%%%%%%%%%%%%%%%%%%%%%%%%%%%%%%

%%%%%%%%%%%%%%%%%%%%%%%%%%%%%%%%%%%%%%%%%%%%%%%%%%%%%%%%%%%%%%%%%%%%%%%%%%%%%%%%%%%%%%%%%%%%%%%%%%
%%%%%%%%%%%%%%%%%%%%%%%%%%%%%%%%%%%%%%%%%%%%%%%%%%%%%%%%%

%%%%%%%%%%%%%%%%%%%%%%%%%%%%%%%%%%%%%%%%%%%%%%%%%%%%%%%%%%%%%%%%%%%%%%%%%%%%%%%%%%%%%%%%%%%%%%%%%%
%%%%%%%%%%%%%%%%%%%%%%%%%%%%%%%%%%%%%%%%%%%%%%%%%%%%%%%%%

%%%%%%%%%%%%%%%%%%%%%%%%%%%%%%%%%%%%%%%%%%%%%%%%%%%%%%%%%%%%%%%%%%%%%%%%%%%%%%%%%%%%%%%%%%%%%%%%%%
%%%%%%%%%%%%%%%%%%%%%%%%%%%%%%%%%%%%%%%%%%%%%%%%%%%%%%%%%

%\scriptsize
%\bibliographystyle{IEEEbib}
%\setlength\itemsep{0em}
%\bibliography{name}
%\bibliography{strings,refs}

%\end{document}

\end{document}